\documentclass[12pt,a4paper]{article}
\usepackage[margin=2cm,headheight=14pt]{geometry}
\usepackage[T1]{fontenc}
\usepackage[utf8]{inputenc}
\usepackage{amsmath,amssymb,mathtools}
\usepackage{newtxtext,newtxmath}
\usepackage[scaled=0.92]{helvet}
\usepackage{textcomp}
\usepackage{microtype}
\usepackage{bm}
\usepackage{booktabs}
\usepackage{multirow}
\usepackage{array}
\usepackage{siunitx}
\usepackage{threeparttable}
\usepackage{tabularx}
\usepackage{longtable}
\usepackage{pdflscape}
\usepackage{graphicx}
\usepackage{subcaption}
\usepackage{float}
\usepackage{afterpage}
\usepackage[section]{placeins}
\usepackage{enumitem}
\usepackage{setspace}
\usepackage{changepage}
\setlist{nosep,topsep=5pt,itemsep=2pt,parsep=1pt,leftmargin=1.8em}
\usepackage{caption}
\usepackage{titlesec}
\usepackage[backend=biber,style=authoryear,natbib=true,maxcitenames=2,mincitenames=1,sortcites]{biblatex}
\newbibmacro*{cite:complete}{%
  \iffieldundef{shorthand}
    {\ifboolexpr{test {\ifnameundef{labelname}}
                 or test {\iffieldundef{labelyear}}}
       {\iffieldundef{label}
          {\printfield{labeltitle}}
          {\printfield{label}}%
        \setunit{\printdelim{nonameyeardelim}}}
       {\printnames{labelname}%
        \setunit{\printdelim{nameyeardelim}}}%
     \printlabeldateextra}
    {\printfield{shorthand}}}
\DeclareCiteCommand{\parencite}
  {\usebibmacro{prenote}\bibopenparen}
  {\usebibmacro{citeindex}\printtext[bibhyperref]{\usebibmacro{cite:complete}}}
  {\multicitedelim}
  {\usebibmacro{postnote}\bibcloseparen}
\DeclareCiteCommand*{\parencite}
  {\usebibmacro{prenote}\bibopenparen}
  {\usebibmacro{citeindex}\printtext[bibhyperref]{\usebibmacro{cite:complete}}}
  {\multicitedelim}
  {\usebibmacro{postnote}\bibcloseparen}
\usepackage{xcolor}
\definecolor{houseblue}{HTML}{1A73E8}
\usepackage{hyperref}
\hypersetup{colorlinks=true,linkcolor=houseblue,citecolor=houseblue,urlcolor=houseblue}
\usepackage{cleveref}
\titleformat{\section}{\sffamily\bfseries\Large}{\thesection}{0.75em}{}
\titleformat{\subsection}{\sffamily\bfseries\large}{\thesubsection}{0.75em}{}
\titleformat{\subsubsection}{\sffamily\bfseries\normalsize}{\thesubsubsection}{0.75em}{}
\titleformat{\paragraph}[runin]{\sffamily\bfseries\normalsize}{}{0pt}{}
\titlespacing*{\section}{0pt}{2.0ex plus .5ex minus .2ex}{1.0ex plus .2ex}
\titlespacing*{\subsection}{0pt}{1.5ex plus .4ex minus .2ex}{0.8ex plus .2ex}
\titlespacing*{\subsubsection}{0pt}{1.2ex plus .3ex minus .2ex}{0.6ex plus .2ex}
\titlespacing*{\paragraph}{0pt}{1.0ex plus .2ex minus .1ex}{0.7em}
\usepackage{fancyhdr}
\fancypagestyle{plain}{\fancyhf{}\fancyfoot[C]{\thepage}}

\renewenvironment{abstract}{\par\begingroup\small\noindent\rule{\textwidth}{0.6pt}\par\begin{adjustwidth}{1.5em}{1.5em}\noindent\textsf{\bfseries Abstract.}\space}{\par\end{adjustwidth}\noindent\rule{\textwidth}{0.6pt}\par\endgroup}
\renewcommand{\footnoterule}{\kern-3pt\hrule width 0.35\textwidth height 0.4pt\kern 2.6pt}

\crefname{equation}{Eq.}{Eqs.}
\Crefname{equation}{Equation}{Equations}
\newcommand{\bps}{\mathrm{bps}}
\title{\textbf{HARN: Hierarchical Associative Resonance Network for Event-Driven Multi-Timeframe Forecasting}}

\author{
Nabeel Ahmad Saidd\\[2pt]
\normalsize Dr. A. P. J. Abdul Kalam Technical University, Lucknow, India\\
\normalsize \href{mailto:nabeelahmadsaidd@gmail.com}{{nabeelahmadsaidd@gmail.com}}
}

\date{}
\begin{document}
\maketitle

\begin{abstract}
Live financial forecasting must synchronize asynchronous bar completions across timeframes without using stale or unavailable information. We introduce HARN (Hierarchical Associative Resonance Network), an event-driven architecture that anchors each event to the finest configured timeframe and maintains persistent states and matrix-valued associative memories across temporal levels. Its timestamp-masked transition updates a level only when its completed-bar timestamp matches the current anchor, preserving inactive representations. HARN integrates causal multi-scale encoding, residual-driven gated associative writes, all-to-all cross-level resonance, and ordered bottom-up evidence aggregation to produce forecasts at each temporal level. Predictions are expressed as basis-point changes and reconstructed on the price scale for evaluation. We evaluate HARN on four assets spanning equities, foreign exchange, and gold (AAPL, EURUSD, USDCHF, and XAUUSD), using chronological 70/15/15 splits, 32-bar lookbacks, one-step horizons, and three random seeds. HARN obtains lower anchor-timeframe test mean absolute errors (MAEs) than the evaluated single-timeframe PatchTST and TimeXer baselines on all four assets. HARN's MAEs are 0.508, 0.000306, 0.000263, and 2.552, respectively; PatchTST records 2.074, 0.000759, 0.000638, and 9.563, while TimeXer records 2.882, 0.010838, 0.014144, and 15.297. These descriptive comparisons do not isolate architectural effects from differences in inputs, target formulations, or training procedures. Single-run component ablations on AAPL and EURUSD show small, non-monotonic differences. The results demonstrate HARN's feasibility for asynchronous multi-timeframe forecasting, while motivating controlled comparisons and broader ablations to establish its relative advantages and component contributions.
\end{abstract}


\clearpage
\section{Introduction}

\label{sec:intro}

A forecaster operating on live financial data must account for a temporal structure that standard sequence-modeling benchmarks often abstract away: observations at different resolutions do not become available simultaneously. A five-minute bar may close while the corresponding hourly or four-hour bar is still forming. Consequently, the \emph{event time}, when a forecast is requested, and the \emph{representation time}, the timestamp of the latest completed bar at each timeframe, need not coincide above the finest resolution. A valid forecasting system must distinguish newly available information from representations that remain unchanged. Reading a still-forming higher-timeframe bar, or refreshing a coarse representation when no new bar has closed, changes the information set available to the model and can introduce look-ahead leakage \parencite{kaufman2012leakage,tashman2000out,bergmeir2018validity}.

This issue is especially important in financial time series, where volatility clustering, changing conditional dependence, and heavy-tailed returns contribute to a non-stationary statistical environment \parencite{cont2001empirical,engle1982arch,bollerslev1986generalized}. Regime-switching and state-space models provide established representations of evolving market dynamics \parencite{hamilton1994time,tsay2010analysis}, whereas modern neural methods learn temporal representations directly from historical observations. Regardless of the modeling paradigm, a live multi-timeframe system must first specify which information is available at each event and when each temporal representation may change.

Most neural forecasting architectures map fixed windows to future values. Recurrent models maintain sequential state \parencite{hochreiter1997long,voelker2019legendre}, and state-space models provide alternative mechanisms for persistent temporal representation \parencite{gu2022efficient,smith2023simplified,gu2023mamba}. These approaches generally advance their state whenever the input sequence advances; they do not, by themselves, represent a hierarchy in which an hourly representation remains unchanged across several intervening five-minute events. Transformers provide flexible interactions across historical windows \parencite{vaswani2017attention}, but attention over resampled or regularly indexed contexts does not specify when asynchronous representations should be refreshed. Resampling, concatenation, or forward filling places coarse information on a common grid, yet does not distinguish a newly completed bar from a carried-forward value. Training an independent model for each timeframe avoids this ambiguity but forfeits an integrated mechanism for cross-scale interaction.

These observations motivate an event-driven formulation in which each timeframe maintains its own representation and updates it only when new information at that resolution becomes available. Such a formulation requires three ingredients: causal observation alignment, persistent state between updates, and an explicit rule for communication across temporal levels. Prior work provides relevant building blocks. Financial econometric models describe evolving conditional dependence and volatility \parencite{hamilton1994time,tsay2010analysis}; neural forecasting architectures exploit long temporal contexts through attention, patching, decomposition, and frequency-aware representations \parencite{zhou2021informer,wu2021autoformer,zhou2022fedformer,nie2023patchtst,woo2023timesnet,wang2024timexer}; and associative-memory approaches provide learned key--value storage and retrieval \parencite{ba2020using,schlag2021linear,hopfield1982neural,ramsauer2021hopfield}. These components motivate the design developed here; we do not claim an exhaustive distinction from all prior multi-timeframe or event-driven systems.

We introduce HARN, a Hierarchical Associative Resonance Network for event-driven multi-timeframe forecasting. HARN treats the fastest configured timeframe as an anchor event stream and maintains a persistent state for every configured temporal level. Each level has a causal multi-scale encoder, a gated associative memory, and a forecasting head. At an anchor event, the model receives the most recent completed window available at each level. A level's state and memory are updated only when its completed-bar indicator fires; otherwise, its previous representation is carried forward. States available at the same event interact through cross-level resonance, while a directional bottom-up readout passes information from lower to higher temporal levels. Forecasting is performed in basis-point change space, and forecasts are reconstructed to the original price scale for evaluation. HARN is thus an event-driven state-transition system, of which the neural components described here are one parameterization.

The central contribution is an explicit formulation and implementation of persistent, asynchronously updated multi-timeframe representations, not a claim of universal predictive superiority. We evaluate HARN on four assets (AAPL, EURUSD, USDCHF, and XAUUSD) using the accompanying preprocessing, training, and evaluation pipeline. The analysis reports reconstructed-price forecasts, comparisons with single-timeframe PatchTST and TimeXer baselines, and component ablations. Because the baselines are trained only on the anchor timeframe, their comparison with HARN is descriptive rather than information-matched; differences in input information, target formulation, and training procedure all affect its interpretation. A code-level audit further examines consistency with the event-level causal protocol. This audit is a static code review, not an empirical leakage test.

The contributions of this paper are as follows:

\begin{enumerate}[leftmargin=*]

\item We formalize an event-driven multi-timeframe forecasting protocol based on completed-bar alignment, anchor events, persistent per-level state, and asynchronous update indicators, including its batched training formulation.

\item We develop HARN as a neural parameterization of this protocol, combining causal multi-scale encoding, gated associative memory, cross-level resonance, and directional bottom-up evidence aggregation.

\item We report the forecasting and ablation results with explicit experimental scope and provenance, distinguishing descriptive baseline comparisons from component-level observations.

\item We provide a code-level causal-consistency audit and document the available preprocessing, configuration, checkpoint, and per-seed result artifacts to support reproducibility.

\end{enumerate}

The remainder of the paper is organized as follows. \Cref{sec:related} reviews related work. \Cref{sec:problem,sec:architecture,sec:streaming} formalize the forecasting protocol, HARN architecture, and streaming and training procedures. \Cref{sec:protocol,sec:results,sec:ablation} describe the experimental setup, forecasting results, and ablation analysis. \Cref{sec:audit} reports the implementation audit, while \Cref{sec:discussion,sec:conclusion} discuss limitations and identify directions for further validation.


\section{Related Work}
\label{sec:related}

Financial forecasting begins in a setting where the statistical properties of the observed process are neither fixed nor necessarily stable. Prices, returns, volatility, and trading activity exhibit different forms of dependence, and the efficient-markets literature sets a demanding benchmark: persistent predictability must be demonstrated empirically rather than assumed \parencite{fama1970efficient}. Financial returns exhibit heavy tails, volatility clustering, and changing dependence structures \parencite{cont2001empirical}. ARCH and GARCH models formalize time-varying conditional variance \parencite{engle1982arch,bollerslev1986generalized}, while state-space and regime-switching formulations allow the underlying data-generating process to evolve \parencite{hamilton1994time,tsay2010analysis}. Under these conditions, evaluation protocol is as important as model capacity. A forecasting system can exploit scale artifacts, normalization statistics, temporal dependence, or a favorable subperiod without learning a relationship that remains available at deployment. Random or otherwise inappropriate validation schemes can therefore misrepresent forecasting performance in non-stationary settings \parencite{bergmeir2018validity}.

This concern extends beyond the train--validation--test split. Temporal leakage may arise through feature construction, normalization, resampling, target alignment, hyperparameter selection, or inadvertent reuse of future information \parencite{kaufman2012leakage}. Forecast evaluation must respect both dependence among forecast errors and the information set available at each forecast origin \parencite{tashman2000out,bergmeir2018validity}. No single scale-free error measure is uniformly appropriate across forecasting problems \parencite{hyndman2006another}. Comparative evaluation may therefore require tests based on loss differentials across dependent forecast origins, such as the Diebold--Mariano framework \parencite{diebold1995comparing}, or conditional predictive-ability tests when relative performance depends on the information set \parencite{giacomini2006tests}. For probabilistic forecasts, proper scoring rules provide the corresponding basis for evaluating predictive distributions \parencite{gneiting2007strictly}. This work adopts the implementation-level implications of that literature: chronological separation, training-only normalization, completed-bar alignment, target isolation, causal feature construction, explicit state-reset boundaries, and validation-only checkpoint selection. The code-level audit in \Cref{sec:audit} examines these conditions. The forecasting target is a basis-point change rather than an absolute price level, following the broader practice of modeling returns and related transformations \parencite{tsay2010analysis,cont2001empirical}. This is a design rationale, not an experimentally established advantage, because the experiments do not isolate target representation.

Within this setting, neural sequence models have substantially expanded the class of temporal dependencies that can be represented. DeepAR learns shared recurrent probabilistic models across related series \parencite{salinas2020deepar}; N-BEATS constructs forecasts through backcast and forecast transformations \parencite{oreshkin2020nbeats}; and Temporal Fusion Transformers combine recurrent processing, attention, and variable selection for multi-horizon forecasting \parencite{lim2021temporal}. The Transformer introduced global self-attention as an alternative to recurrence \parencite{vaswani2017attention}, and subsequent time-series architectures explored sparse attention, decomposition, frequency-domain representations, patch-based tokenization, temporal reshaping, and cross-attention. Informer reduces attention cost through sparse attention and distillation \parencite{zhou2021informer}; Autoformer and FEDformer introduce decomposition-based approaches with autocorrelation or frequency-enhanced attention \parencite{wu2021autoformer,zhou2022fedformer}; PatchTST represents time-series segments as patches \parencite{nie2023patchtst}; TimesNet maps temporal variation into a two-dimensional representation \parencite{woo2023timesnet}; and TimeXer uses cross-attention to incorporate exogenous variables \parencite{wang2024timexer}. Controlled multi-asset benchmarking further illustrates the value of comparing such architectures under a common forecasting setup \parencite{Saidd2026Benchmark}.

These architectures are strong alternatives for temporal representation learning, but their usual forecasting protocol assumes a preconstructed sequence or covariate panel whose indices define temporal progression. In a multi-timeframe financial stream, representations become informative at different event times: a coarse bar may remain unchanged across several fine-scale events, whereas a newly completed fine-scale bar may require an immediate update. Treating these observations as a regular sequence therefore requires an explicit rule for when each representation is refreshed and how an inactive representation is carried forward. HARN makes this rule part of the model specification rather than an external preprocessing assumption. Each timeframe has a persistent state and an update indicator; when the indicator is inactive, the corresponding state and associative memory are preserved rather than overwritten (\cref{eq:maskedtransition}). Cross-level interaction then proceeds through resonance (\cref{eq:resonance}) and a defined lower-to-higher evidence pathway (\cref{eq:evidence}). This distinction concerns forecasting protocol and state-transition semantics, not a rejection of attention-based modeling: HARN itself uses attention-like interactions across levels and in its evidence readout.

Persistent state is also central to recurrent and state-space approaches. LSTM introduced gated memory mechanisms that substantially improved the representation of long-range dependencies \parencite{hochreiter1997long}. Legendre Memory Units encode long histories through structured continuous-time dynamics \parencite{voelker2019legendre}, while structured state-space models provide explicit state-transition formulations with efficient long-sequence computation \parencite{gu2022efficient,smith2023simplified,gu2023mamba}. Selective state-space models, including Mamba, further condition state transitions on the input itself \parencite{gu2023mamba}. HARN is related to this family through its compact persistent states, but it makes the update event an explicit component of the transition rule. Conventional recurrent formulations ordinarily advance a state at every sequence index; HARN instead applies a per-timeframe mask, allowing one level to remain unchanged while other levels receive new observations (\cref{eq:maskedtransition}). Its state is therefore a coupled collection of level-specific recurrences operating under an event-conditioned schedule. Recurrent and state-space models are relevant architectural alternatives, but they are not tested controls for the event-conditioned transition mechanism studied here.

A second relevant line of work concerns associative memory. Hopfield networks established content-addressable memory through an energy-based attractor formulation \parencite{hopfield1982neural}. Fast-weight methods subsequently introduced temporary key--value structures that can be updated to encode recent context \parencite{ba2020using}, and linear Transformers admit an interpretation in terms of fast-weight programming and associative updates \parencite{schlag2021linear}. Modern Hopfield networks connect associative retrieval with attention-like mechanisms and substantially extend the representational capacity of the original formulation \parencite{ramsauer2021hopfield}. HARN uses a narrower mechanism designed for its event-driven setting. Its encoder output and cross-scale context yield a normalized key, a value, and a normalized query; the difference between the current value and the retrieved memory content forms a retrieval residual, referred to as ``surprise'' in \Cref{sec:memory}, which controls a bounded outer-product write (\cref{eq:memory}). The mechanism thus separates retrieval from writing rather than treating memory solely as another recurrent hidden-state transformation. The A1 ablation replaces this associative cell with a gated recurrent-style update; because that comparison is not parameter-count matched, it does not isolate memory capacity as a single causal factor (\Cref{sec:ablation}).

The temporal hierarchy considered by HARN is also related to hierarchical forecasting, although the objectives differ. Hierarchical forecasting concerns forecasts defined at multiple aggregation levels and, in many formulations, their reconciliation into a coherent collection; optimal-combination methods construct forecasts that respect aggregation relationships \parencite{hyndman2011optimal}. Neural architectures such as N-BEATS and Temporal Fusion Transformers learn multi-scale or multi-horizon representations \parencite{oreshkin2020nbeats,lim2021temporal}, and decomposition-based Transformer models similarly separate temporal variation at different scales \parencite{wu2021autoformer,zhou2022fedformer}. HARN uses the term ``hierarchical'' in a related but distinct sense. Its hierarchy consists of timeframes whose bars complete at different rates, and it determines when a persistent representation is eligible for a state transition rather than imposing an output-side aggregation constraint. A coarse representation is not merely a lower-frequency forecasting head: it retains its own state while receiving contemporaneous cross-level context and lower-level evidence. Conversely, repeating the most recent coarse value does not record whether a new coarse bar has actually completed, so HARN treats completion status as explicit information through the update indicators. The reported experiments do not compare this mechanism with reconciled hierarchical forecasting methods or with explicitly matched multi-scale neural baselines.

Irregular and event-driven sequence models offer another point of comparison. Latent ODEs model hidden trajectories at arbitrary observation times by integrating a learned vector field \parencite{rubanova2019latent}, and Neural Controlled Differential Equations treat an observed path as a control signal driving continuous hidden dynamics \parencite{kidger2020neural}. These approaches handle observations that are not regularly spaced in a principled way. The multi-timeframe financial setting considered here is more structured: each timeframe has a known bar-completion rule, observations have ordered timestamps, each input window contains completed bars, and the update status of a level follows from timestamp equality. HARN consequently uses discrete event ordering and state retention rather than inferring a continuous hidden trajectory between observations. This makes the causal contract directly inspectable, but it also makes the implementation dependent on the specified timestamp and bar-construction conventions, which are examined in \Cref{sec:audit}.

Taken together, these literatures supply the components from which HARN is constructed. Financial econometrics motivates careful treatment of non-stationarity and information availability \parencite{hamilton1994time,cont2001empirical}; recurrent and state-space models provide mechanisms for compact persistent state \parencite{hochreiter1997long,gu2023mamba}; Transformers provide flexible cross-temporal interaction \parencite{vaswani2017attention,nie2023patchtst}; associative-memory methods provide content-addressable retrieval and update \parencite{hopfield1982neural,schlag2021linear}; hierarchical forecasting addresses relationships across temporal aggregation levels \parencite{hyndman2011optimal}; irregular-time models offer alternatives for asynchronous observations \parencite{rubanova2019latent,kidger2020neural}; and leakage and forecast-evaluation research establishes the need to align computation with the information set available at each forecast origin \parencite{kaufman2012leakage,bergmeir2018validity,diebold1995comparing}. The more specific gap addressed by HARN lies in how these ideas are combined for a stream of completed financial bars observed at multiple resolutions.

To our knowledge, the literature reviewed here does not provide a single established protocol that simultaneously specifies per-timeframe completion events, persistence of inactive higher-level states, within-event cross-level communication, and compact associative state updates under an explicit causal ordering. We frame this as a gap in protocol and state-transition specification, not as a claim that prior methods lack any individual component. HARN addresses the combination through an event-conditioned state transition, a bounded residual-driven associative write (\cref{eq:memory}), cross-level resonance around selective updates (\cref{eq:resonance}), and a lower-to-higher evidence pathway with an explicit history order (\cref{eq:evidence}). The result is a testable specification for event-driven multi-timeframe state rather than a collection of isolated architectural claims. Accordingly, the empirical evidence is interpreted within the scope of this specification and its single implementation, not as evidence of superiority over the broader forecasting literature (\Cref{sec:protocol,sec:results,sec:discussion}).

\section{Problem Formulation: Event-Driven Multi-Timeframe Forecasting}
\label{sec:problem}
This section formalizes the protocol that \Cref{sec:related} identified as under-specified. It fixes the notation used throughout and states what a causally valid multi-timeframe forecast may and may not depend on; this is the standard against which \Cref{sec:audit} checks the implementation.

Let the levels be ordered from finest to coarsest as $T_1,\ldots,T_L$. A completed bar at level $k$ is $x_t^k=(o_t^k,h_t^k,\ell_t^k,c_t^k,v_t^k)\in\mathbb{R}^5$, the usual open--high--low--close--volume representation of a fixed time interval. \emph{Event time} $\tau$ is the sequence of anchor timestamps at which the system is queried for a forecast, and \emph{representation time} is the timestamp of the most recently completed bar available at each level. The two clocks coincide at the finest level by construction, since the anchor stream is defined by that level's bar completions, but they need not coincide at coarser levels. We write $r_\tau^k,M_\tau^k$ for the persistent state immediately before processing event $\tau$, and $r_{\tau+1}^k,M_{\tau+1}^k$ for the state stored immediately after that event and used by the forecast emitted at $\tau$. The subscript $\tau+1$ thus denotes the post-event state; it does not imply that the forecast waits for the next event.

\paragraph{Update indicator.}
Events are indexed by the finest level. At anchor time $\tau$, preprocessing selects the latest level-$k$ bar whose timestamp is no later than $\tau$ and sets
\begin{equation}
 u_\tau^k=\mathbf{1}\{\text{selected timestamp equals }\tau\}.
 \label{eq:update}
\end{equation}
The no-later-than rule prevents a future bar from entering an input window, while the equality test determines separately whether the persistent level state may be overwritten. Thus, $u_\tau^k=0$ means that the selected higher-level window remains available as context, but no new completed bar closed exactly at $\tau$; the level contributes its most recent valid representation and its persistent state is preserved. Exact timestamp equality is an implementation assumption, not a claim that every market feed exposes perfectly aligned timestamps; \Cref{sec:audit} revisits this assumption.

\paragraph{Persistent state and masked transition.}
For each level, HARN maintains a state $r_\tau^k\in\mathbb{R}^{d_s}$ and a matrix memory $M_\tau^k\in\mathbb{R}^{d_s\times d_m}$. The state is a compact vector summary analogous to a recurrent hidden state, whereas the memory is a matrix-valued associative store that supports content-based retrieval, as in fast-weight and Hopfield-style memories \parencite{hopfield1982neural,ba2020using,ramsauer2021hopfield}. Let $\tilde r_{\tau+1}^k$ and $\tilde M_{\tau+1}^k$ denote candidate updates computed from the current payload: the state and memory that the level would adopt if it were active at this event. For a single stream and $k>1$, the persistent transition is
\begin{align}
 r_{\tau+1}^k&=(1-u_\tau^k)\,r_\tau^k+u_\tau^k\,\tilde r_{\tau+1}^k,\notag\\
 M_{\tau+1}^k&=(1-u_\tau^k)\,M_\tau^k+u_\tau^k\,\tilde M_{\tau+1}^k,
 \label{eq:maskedtransition}
\end{align}
where $u_\tau^k\in\{0,1\}$ is a scalar that multiplies the whole state or memory. The base level, whose bar closes at every anchor event, has $u_\tau^1\equiv1$ and is updated without masking. Because the indicator is binary rather than learned, this convex combination expresses ``no new information at this level, so do not change its state'' as an exact identity transition.

\paragraph{Batched form used in training.}
Training and evaluation process $B$ contiguous streams in parallel (\Cref{sec:streaming}). Let $b\in\{1,\ldots,B\}$ index streams, and let $u_{b,\tau}^k$ be the indicator in \cref{eq:update} for stream $b$ at its current event. We use $\tau$ for the step index and suppress each stream's separate timestamp. The indicators at one step form the mask vector
\begin{equation}
 m_\tau^k=\bigl(u_{1,\tau}^k,\ldots,u_{B,\tau}^k\bigr)^{\!\top}\in\{0,1\}^B .
 \label{eq:batchedmask}
\end{equation}
With stacked states $R_\tau^k\in\mathbb{R}^{B\times d_s}$ and memories $\mathcal{M}_\tau^k\in\mathbb{R}^{B\times d_s\times d_m}$, the implemented update is
\begin{equation}
\begin{aligned}
 R_{\tau+1}^k&=(1-m_\tau^k)\odot_B R_\tau^k+m_\tau^k\odot_B\tilde R_{\tau+1}^k,\\
 \mathcal{M}_{\tau+1}^k&=(1-m_\tau^k)\odot_B\mathcal{M}_\tau^k+m_\tau^k\odot_B\tilde{\mathcal{M}}_{\tau+1}^k,
\end{aligned}
 \label{eq:batchedtransition}
\end{equation}
where $\odot_B$ multiplies the $b$-th slice of a tensor by the $b$-th entry of the mask, broadcasting over the trailing dimensions. Slice $b$ of \cref{eq:batchedtransition} is exactly \cref{eq:maskedtransition} for stream $b$, so the scalar definition and the batched implementation coincide. The candidates $\tilde R$ and $\tilde{\mathcal{M}}$ are computed for the entire batch regardless of the mask; masking then selects between the two already-computed branches.

\paragraph{Causal requirement.}
Let $\mathcal{I}_\tau=\{x_s^j:\ s\le\tau,\ 1\le j\le L\}$ denote the completed bars at all levels with timestamps no later than $\tau$. A causally valid forecaster must satisfy
\begin{equation}
 \hat y_{\tau+1}^k=f_k\bigl(\mathcal{I}_\tau;\,r_0^{1:L},M_0^{1:L}\bigr),
 \label{eq:causal}
\end{equation}
where $(r_0^{1:L},M_0^{1:L})$ is the reset initial state and the dependence on $\mathcal{I}_\tau$ is mediated by the persistent states obtained by applying the transition to completed bars in order. The requirement is event-level: no bar completed after $\tau$, and no state derived from such a bar, may influence $\hat y_{\tau+1}^k$. It does not require the level-$k$ update at event $\tau$ to be independent of computations at other levels during that event. HARN's bottom-up evidence path allows a higher-level update at $\tau$ to read the lower-level state updated earlier in the same event, and the forecasting head consumes the post-update states $r_{\tau+1}^k$ (\Cref{sec:architecture}). HARN therefore satisfies \cref{eq:causal} at event granularity, but it is not past-event-only within an event. This qualification recurs as entry~9 of the audit (\Cref{tab:audit}). HARN does enforce a stronger local property for its convolutions and shifted features (\Cref{sec:architecture}).

\paragraph{Implementation notes and data handling.}
The implementation computes candidate encodings for every supplied level on each forward call, including inactive levels; masking governs persistent overwrites, not computation skipping. ``Inactive'' therefore means that a level's persistent state is protected, not that its window is untouched. Within an event, lower-level updates finish before their current reads are appended to the higher-level history buffer. Data are cleaned by parsing timestamps, removing invalid rows, retaining the last duplicate, and sorting in ascending order. Splits are chronological 70\%/15\%/15\% \parencite{tashman2000out,bergmeir2018validity}; feature and target statistics are computed on training samples only and reused for validation and test \parencite{kaufman2012leakage}. The resulting formulation is an event-driven state-transition problem, not a regular-grid forecasting problem with additional covariates.

\paragraph{Stationary target.}
Let $P_\tau^k$ be the latest observed close and $P_{\tau+1}^k$ the next completed close at level $k$. The supervised target is the basis-point change
\begin{equation}
 y_{\tau+1}^k=10{,}000\left(\frac{P_{\tau+1}^k}{P_\tau^k}-1\right)\;\bps,
 \label{eq:bps}
\end{equation}
and the prediction is reconstructed as $\hat P_{\tau+1}^k=P_\tau^k(1+\hat y_{\tau+1}^k/10{,}000)$. Modeling a relative change rather than a level is standard practice in financial time-series analysis, since levels differ widely across assets and are highly persistent, whereas relative changes are closer to stationary \parencite{tsay2010analysis,cont2001empirical}. Recent objectives also combine volatility normalization with direction-aware terms \parencite{Saidd2026RNAD}; HARN instead uses unnormalized per-level BPS mean-squared error, as specified in \cref{eq:loss}. The BPS target is dimensionless with respect to price level and is expressed in the units conventionally used for small relative movements. Two consequences matter for interpreting the reported results. First, the reconstruction error is
\begin{equation}
 \hat P_{\tau+1}^k-P_{\tau+1}^k=\frac{P_\tau^k\,(\hat y_{\tau+1}^k-y_{\tau+1}^k)}{10{,}000},
 \label{eq:reconerr}
\end{equation}
so HARN's reported price errors are price-scaled BPS errors anchored at the last observed close, and a zero-BPS forecast reproduces the last-value (persistence) forecast. Second, all reported evaluation metrics are computed on reconstructed absolute price, which reintroduces asset-scale dependence; the BPS target is used for training and reconstruction and is not itself reported as an evaluation metric.


\section{The HARN Architecture}
\label{sec:architecture}
\begin{figure}[H]\centering
\includegraphics[width=0.98\linewidth]{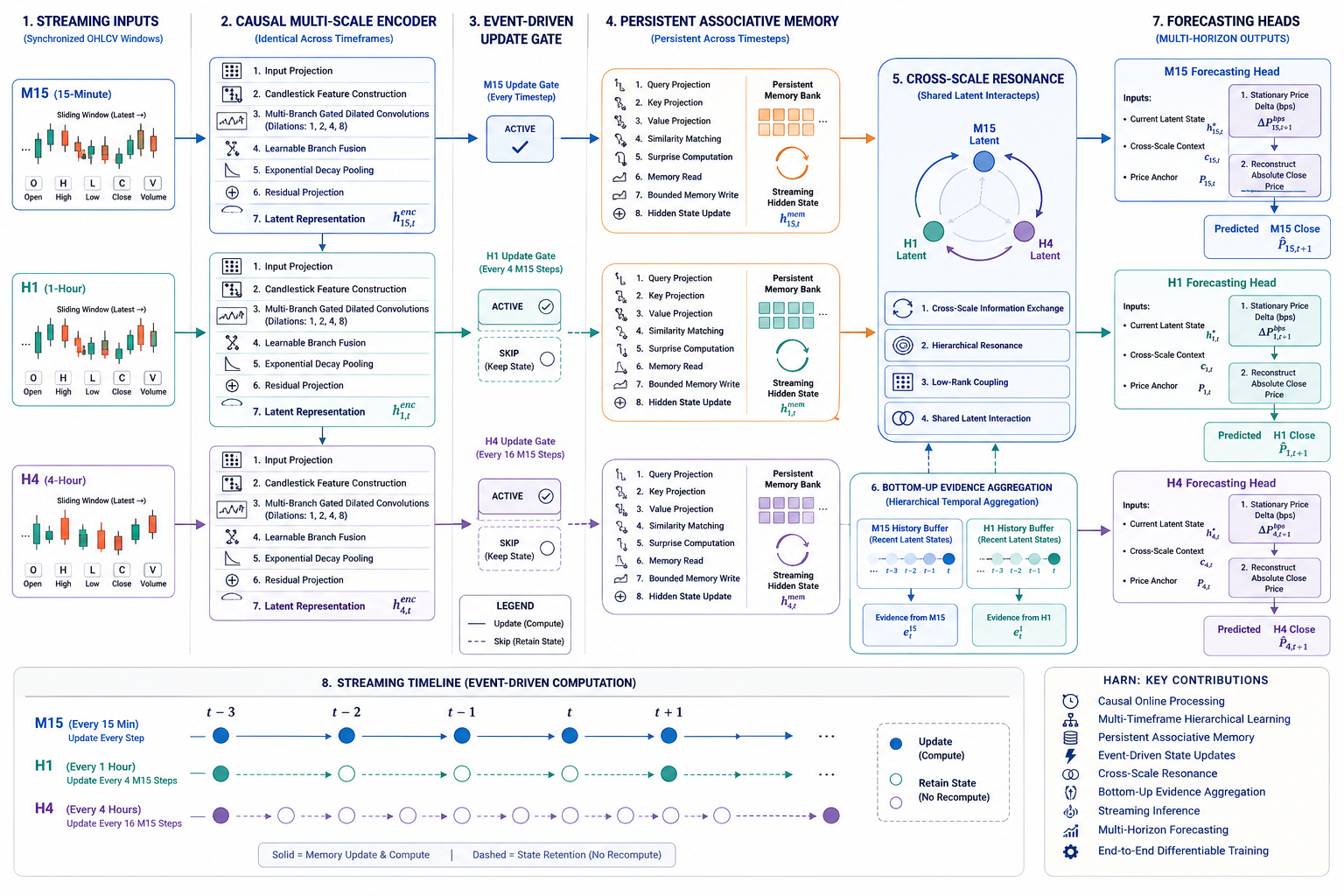}
\caption{HARN's event-driven computational path at one anchor event: per-level causal encoding, prior-state resonance, gated memory candidates, ordered bottom-up evidence, posterior resonance, and per-level forecasting. This conceptual schematic is not an empirical result.}\label{fig:architecture}
\end{figure}

\Cref{fig:architecture} summarizes the computation. At each event, the path is: completed OHLCV windows $\rightarrow$ per-level causal encoders $\rightarrow$ prior-state resonance $\rightarrow$ gated memory candidates $\rightarrow$ ordered bottom-up evidence for higher levels $\rightarrow$ posterior resonance $\rightarrow$ per-level forecasting heads. HARN instantiates separate encoders, memories, histories, and heads for every configured level. The hierarchy is therefore both representational and operational: level identity determines what a state represents and when it may change. This section presents the components in execution order; \Cref{sec:streaming} then assembles them into the complete event protocol. The descriptions below concern implemented computation and are code observations, not experimental findings. In the notation, $t$ indexes positions within an encoder window, whereas the memory, state, and head equations use the event index $\tau$ from \Cref{sec:problem}; $x_t$ denotes the memory input vector in those equations.

\subsection{Causal multi-scale encoder}
\begin{figure}[H]\centering
\includegraphics[width=0.95\linewidth]{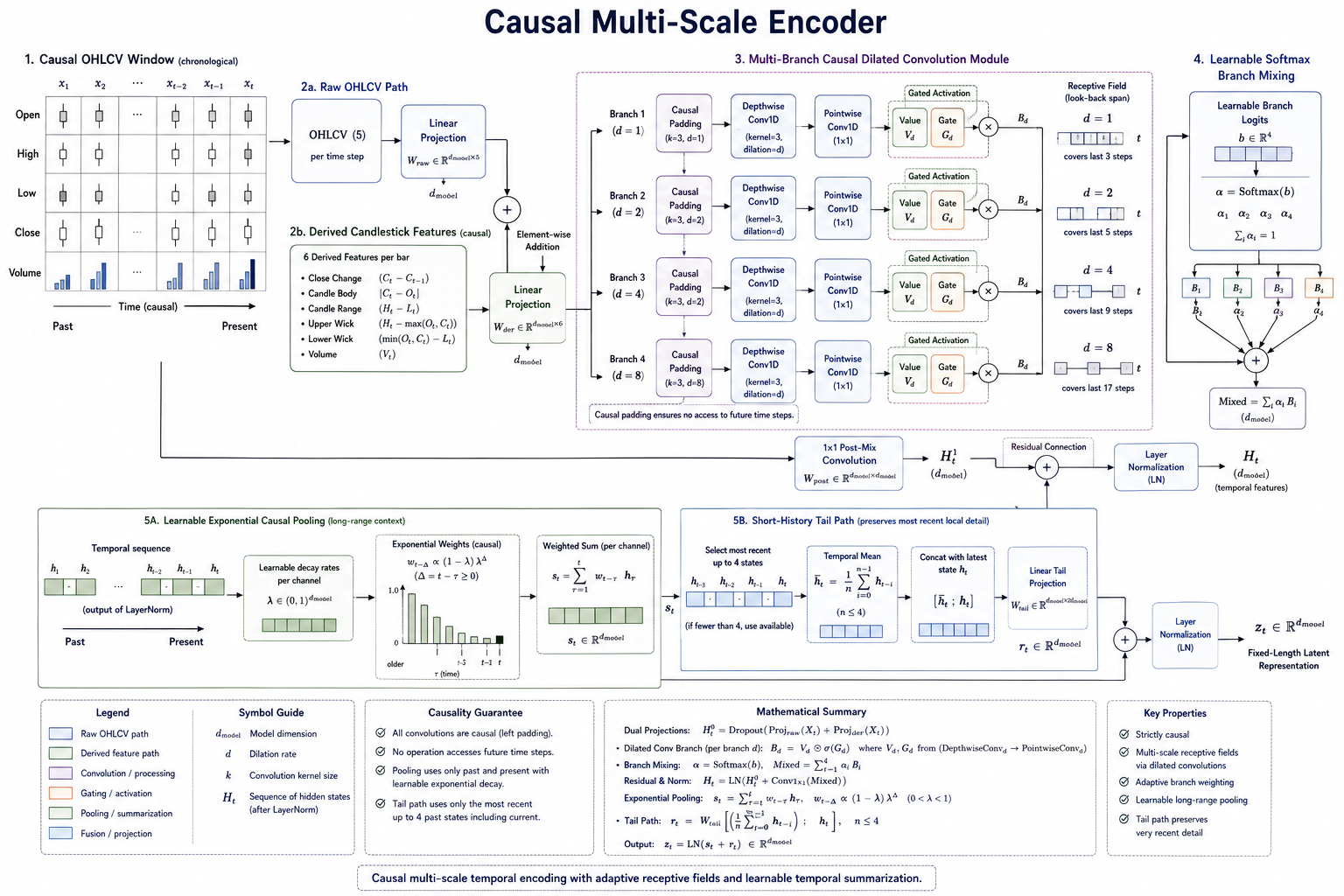}
\caption{Causal multi-scale encoder with left-padded dilated branches, a learned branch mixture, and decayed pooling. This conceptual schematic is not an empirical result.}
\label{fig:encoder}
\end{figure}

The encoder summarizes a completed OHLCV window as a fixed-size representation at multiple temporal scales. It uses only information within that window and never mixes in a later position (\Cref{fig:encoder}). For $X^k\in\mathbb{R}^{B\times T\times5}$, it projects the raw OHLCV features together with six derived features that expose short-horizon price structure to the downstream convolutions:
\begin{align}
 d_t^k&=(c_t-c_{t-1},\,c_t-o_t,\,h_t-\ell_t,\,h_t-\max(o_t,c_t),\,\min(o_t,c_t)-\ell_t,\,v_t),\\
 z_t^k&=W_{\rm raw}x_t^k+W_{\rm derived}d_t^k,
\end{align}
where the previous close is shifted by one position and the first previous close is copied from the first close, avoiding an artificial pre-window observation. The implementation receives normalized OHLCV windows and computes these quantities from the normalized representation. The six derived channels are close change, signed candle body, full candle range, upper wick, lower wick, and the already normalized volume channel; volume is therefore reused rather than transformed a second time. These are candle-geometry and volume features, not a conventional technical-indicator panel, a distinction examined in related indicator-ablation work \parencite{Saidd2026Indicators}. HARN does not isolate the contribution of individual derived channels.

Four depthwise-separable gated convolution branches use kernel size three and dilations $D=\{1,2,4,8\}$. Left-only padding ensures that no output at position $t$ depends on a later position. This follows the causal dilated design used for autoregressive modeling of long sequences, in which exponentially increasing dilation expands the receptive field while the depth grows logarithmically with sequence length \parencite{oord2016wavenet}; depthwise separation further reduces parameters relative to a full convolution \parencite{chollet2017xception}:
\begin{equation}
 C_d=V_d\odot\sigma(G_d),\qquad C=\sum_{d\in D}\operatorname{softmax}(\alpha)_dC_d.
\end{equation}
The gate $\sigma(G_d)$ allows each branch to modulate its contributing channels and positions. The learned softmax mixture over dilations weights temporal scales rather than averaging them uniformly. The mixture then passes through a pointwise convolution, a residual connection, and LayerNorm \parencite{ba2016layer}. A learnable per-channel decay $\lambda_c=\sigma(\phi_c)$ pools the sequence with greater weight on recent positions while retaining a causal contribution from the full window:
\begin{equation}
 p_c=\sum_{t=1}^{T}(1-\lambda_c)\lambda_c^{T-t}C_{t,c}.
 \label{eq:pool}
\end{equation}
The encoder output combines this decayed pool with a tail projection of the last position and a short trailing mean, so that the most recent bar and the immediate local trend are represented directly,
\begin{equation}
 e^k=\operatorname{LN}\left(p^k+W_{\rm tail}\bigl[C_T^k;\operatorname{mean}(C_{T-\min(4,T)+1:T}^k)\bigr]\right)\in\mathbb{R}^{72}
 \label{eq:encoder}
\end{equation}
in the default configuration. This local representation is the only window-derived signal that enters the memory cell; persistence across events enters through the memory cell described next.

\subsection{Gated associative memory with residual-driven writes}
\label{sec:memory}
\begin{figure}[H]\centering
\includegraphics[width=0.86\linewidth]{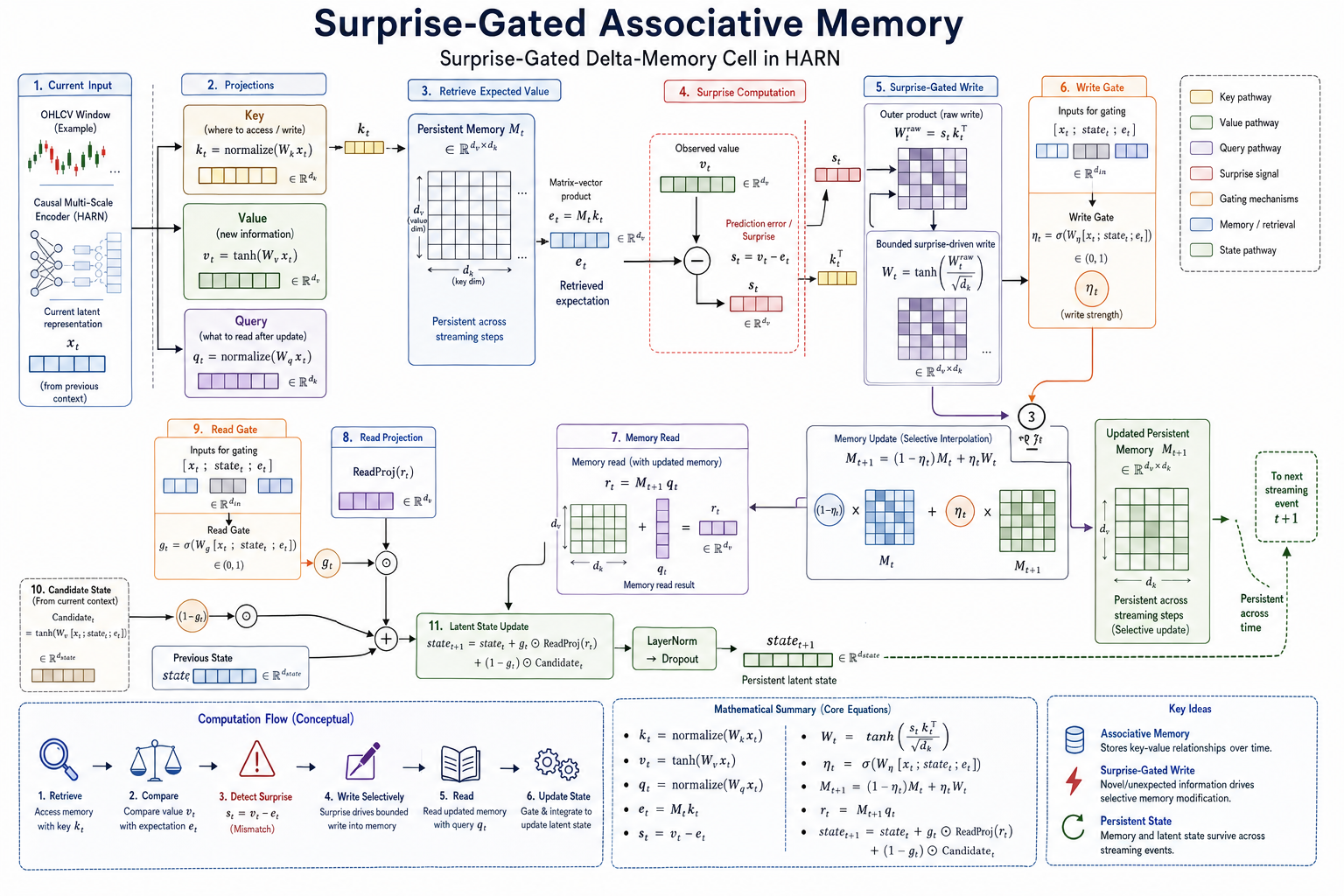}
\caption{Gated associative memory cell showing key, value, and query projections, the retrieval residual $s_t$, and the bounded gated write in \cref{eq:memory}. This conceptual schematic is not an empirical result.}\label{fig:memory}
\end{figure}

The encoder summarizes the current window; the memory cell maintains a persistent, content-addressable representation across events (\Cref{fig:memory}). Its input concatenates the encoder output with prior resonance context and, for levels above the base, bottom-up evidence. The resulting input dimension is 144 at the base level and 216 at higher levels. For normalized input $x_t$, state $r_t$, and memory $M_t$, the cell computes a key, value, and query in the manner of a standard attention or fast-weight mechanism \parencite{schlag2021linear}:
\begin{align}
 k_t&=\operatorname{normalize}(W_kx_t),&v_t&=\tanh(W_vx_t),&q_t&=\operatorname{normalize}(W_qx_t),\\
 \bar v_t&=M_tk_t,&s_t&=v_t-\bar v_t.
\end{align}
The residual $s_t$ measures the discrepancy between the value implied by the current input and the value retrieved from memory for the corresponding key. We use ``surprise'' only as a label: this is an internal retrieval residual, not the downstream forecast error. No reported experiment tests whether it tracks prediction error, novelty, or market events. With $g_t=[x_t;r_t;\bar v_t]$, learned gates control the write and blend rates:
\begin{equation}
 \eta_t=\sigma(W_\eta g_t),\qquad \rho_t=\sigma(W_\rho g_t),
 \qquad \eta_t,\rho_t\in\mathbb{R}^{B\times d_s}.
\end{equation}
The residual enters the \emph{content} of the write, not the gate: $s_t$ is not an explicit input to $\eta_t$ or $\rho_t$. The write is a rank-one, $\tanh$-bounded outer product of the residual and the key, scaled by the square root of the memory-key dimension as in scaled dot-product attention \parencite{vaswani2017attention}:
\begin{align}
 \Omega_t&=\tanh\left(\frac{s_tk_t^\top}{\sqrt{d_m}}\right),\\
 M_{t+1}&=(1-\eta_t)\odot_B M_t+\eta_t\odot_B \Omega_t.
 \label{eq:memory}
\end{align}
Here $M_t,\Omega_t\in\mathbb{R}^{B\times d_s\times d_m}$, while each gate has shape $(B,d_s)$ and is unsqueezed to $(B,d_s,1)$ before multiplication. In this equation, $\odot_B$ therefore denotes row-wise broadcasting over the $d_m$ memory-key columns; it is not a single scalar gate for the whole matrix. The updated memory is then queried with the query vector rather than the key just written, so that retrieval reflects the memory's content as a whole, and the state update combines a memory-derived and a direct pathway,
\begin{align}
 \mu_t&=M_{t+1}q_t,\\
 r_{t+1}&=\operatorname{LN}\left(r_t+\rho_t\odot W_r\mu_t+(1-\rho_t)\odot\tanh(W_cg_t)\right),
 \label{eq:state}
\end{align}
followed by dropout \parencite{srivastava2014dropout}. The default memory shape is $(B,72,18)$. Unlike a standard recurrent cell, which compresses temporal information into a single vector transition, this cell maintains a matrix of key--value associations and exposes the retrieval residual explicitly, in a manner structurally related to Hopfield-style retrieval \parencite{hopfield1982neural,ramsauer2021hopfield}. Relative to generic fast-weight formulations \parencite{ba2020using,schlag2021linear}, the write is bounded by $\tanh$, scaled by the memory-key dimension, and modulated by a state-sized gate. These choices are \emph{intended} to keep the persistent memory numerically stable over long streams, but no stability analysis or capacity claim is made or tested. The A1 ablation (\Cref{sec:ablation}) replaces this cell with a simpler gated recurrent update.

\subsection{Cross-scale communication and forecasting head}
\begin{figure}[!htbp]\centering
\includegraphics[width=0.95\linewidth]{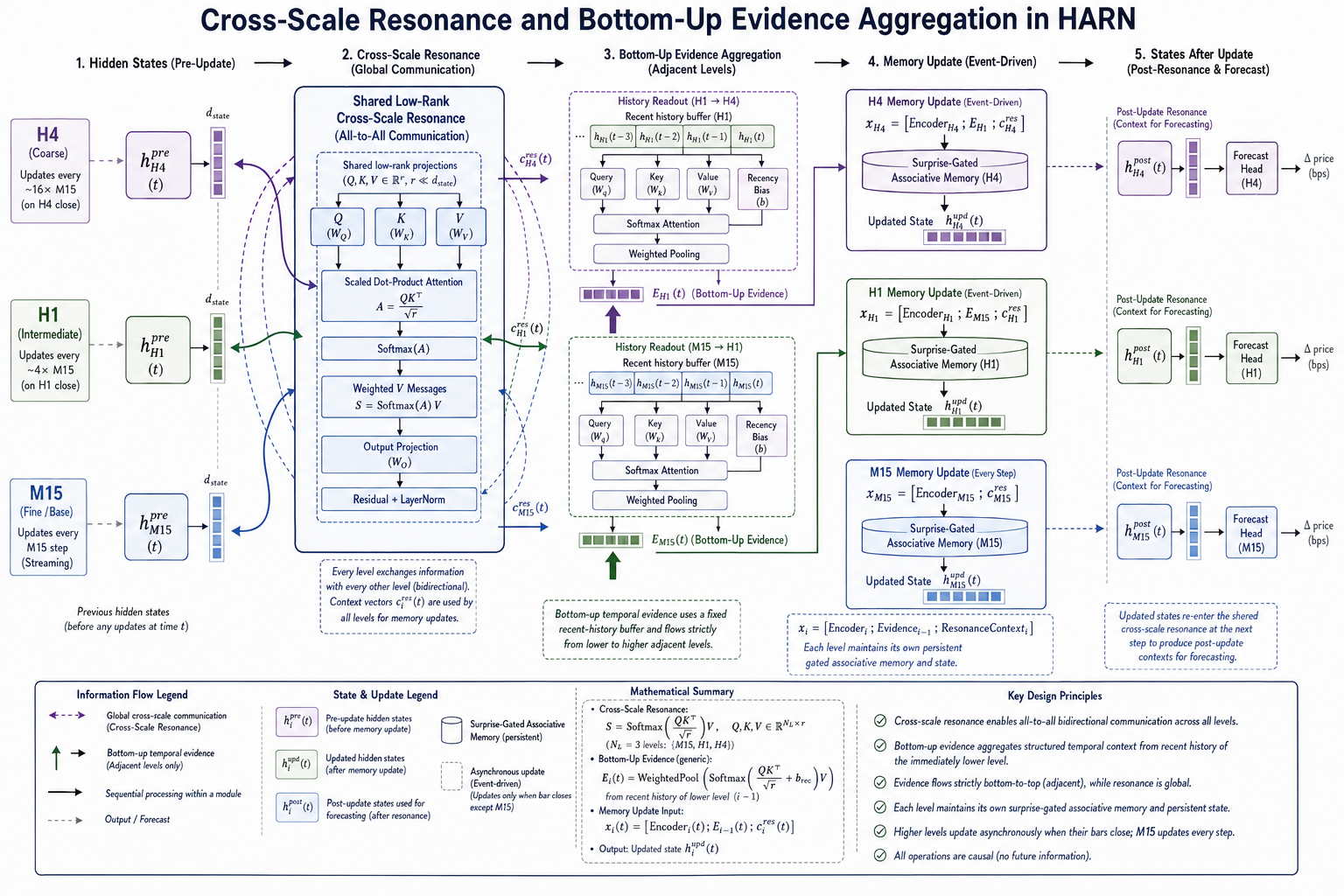}
\caption{HARN's two cross-level communication paths: all-to-all, contemporaneous resonance (\cref{eq:resonance}) and directional, ordered bottom-up evidence (\cref{eq:evidence}). This conceptual schematic is not an empirical result.}\label{fig:crossscale}
\end{figure}

HARN communicates across levels through two paths, resonance and evidence, which differ in topology and timing (\Cref{fig:crossscale}). For resonance, states are stacked as $R\in\mathbb{R}^{B\times L\times d_s}$, and shared projections of rank $d_r=36$ produce an all-to-all attention update along the level axis. The operation follows multi-head self-attention \parencite{vaswani2017attention}, but applies it across levels rather than across time and includes a learned pairwise level-strength bias $S$:
\begin{align}
 Q&=W_qR,&K&=W_kR,&V&=W_vR,\\
 A&=\operatorname{softmax}\left(\frac{QK^\top}{\sqrt{d_r}}+S\right),\\
 R'&=\operatorname{LN}\left(R+\operatorname{Dropout}(W_oAV)\right).
 \label{eq:resonance}
\end{align}
Resonance is applied twice at each event. The first application forms prior context before the memory update, so each level's write can use the other levels' states at the start of the event. The second application operates on the updated states before forecasting. Resonance is all-to-all and contemporaneous: every configured level can attend to every other level at the same event, without a directional constraint.

Bottom-up evidence is instead adjacent-level and directional: a coarser level's update uses information from the immediately finer level. For each adjacent pair, a length-four history buffer stores recent lower-level reads. A learned query attends to projected keys and values with a learnable recency bias. The readout combines the attention-weighted summary with explicit contributions from the most recent read and the buffer mean:
\begin{equation}
 E_i=\operatorname{LN}\left(\operatorname{Attention}(H_i)+0.35\,H_{i,\mathrm{latest}}+0.15\operatorname{mean}(H_i)\right).
 \label{eq:evidence}
\end{equation}
The coefficients $0.35$ and $0.15$ are hand-set implementation constants, not learned parameters. The available materials contain no derivation, tuning record, or sensitivity analysis for them, and it is not documented whether they were tuned or inherited from earlier code. We therefore report them as manually chosen rather than empirically validated; the A3 variant zeros the entire readout and does not probe them individually.

The current lower-level read is appended to $H_i$ before the corresponding higher-level evidence is computed, establishing an ordered same-event path. This is the qualification stated in \Cref{sec:problem}: the path is causal with respect to future events but not restricted to past event states, because it appends the lower level's already completed update. Neither path is computation-free for inactive levels: their encoders and history reads are evaluated, while only persistent state and memory overwrites are masked (\cref{eq:maskedtransition}).

\paragraph{Forecasting head.}
The forecasting head maps the cross-scale-informed state to a one-step BPS forecast. It receives the current state, the posterior resonance context, and six anchor features: one-step and two-step close changes, trailing close deviation, trailing range, trailing body, and mean normalized volume. The concatenated input has dimension $72+72+6=150$ in the default configuration. A two-hidden-layer GELU MLP of width 80 \parencite{hendrycks2016gaussian} with dropout 0.1 \parencite{srivastava2014dropout} is combined with a residual linear path from the current state and anchors, allowing the head to fall back to a linear state-dependent function:
\begin{equation}
 \hat y_{t+1}^k=\operatorname{MLP}\bigl([r_{t+1}^k;R_{t+1}^{\prime k};a_t^k]\bigr)+W_{\rm res}[r_{t+1}^k;a_t^k].
 \label{eq:head}
\end{equation}
Linear layers use Xavier uniform initialization \parencite{glorot2010understanding} and convolution layers use Kaiming uniform initialization \parencite{he2015delving}, with zero biases throughout. Both schemes scale the initial weight variance with the layer's fan, and the choice follows the activation used downstream of each layer type. The head outputs a one-step BPS forecast for each configured level. The absolute-price forecast used for evaluation is obtained only afterwards, by inserting the last observed close into the reconstruction stated after \cref{eq:bps}.


\section{Streaming Semantics and Training}
\label{sec:streaming}
\begin{figure}[http]
\centering
\includegraphics[width=0.96\linewidth]{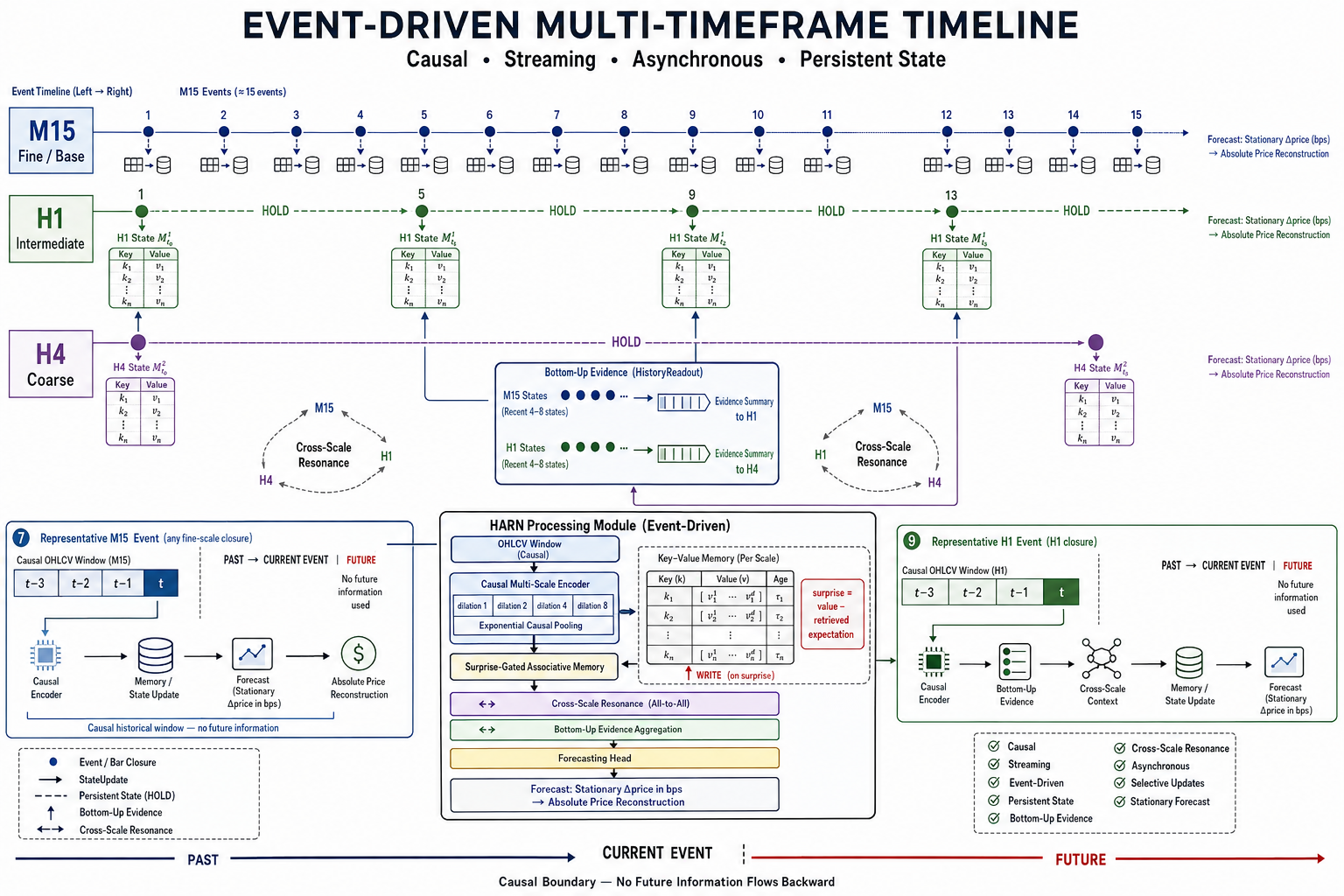}
\caption{Event-driven multi-timeframe timeline. The anchor stream generates forecast events; a coarser representation is updated only when its completed-bar timestamp equals the anchor timestamp and is otherwise carried forward. This conceptual schematic illustrates the protocol and is not an empirical result.}
\label{fig:timeline}
\end{figure}

At each anchor event, the implementation selects completed-bar windows, encodes every supplied window, applies prior-state resonance, constructs anchor features, updates the base level, computes lower-level evidence in order, conditionally updates higher levels, applies posterior resonance, and forecasts at every level. The ordered procedure is restated in \Cref{app:ordering}. Candidate transitions are computed for every level at every event, but for each sample only the candidates whose mask entry in \cref{eq:batchedmask} equals one overwrite persistent state (\cref{eq:batchedtransition}). Because masking follows a batched forward pass, candidate computation covers the whole batch even when only some streams are active. If activity flags are missing, they default to lock-step updates ($m_\tau^k\equiv\mathbf 1$ for every level), the special case in which the mechanism reduces to ordinary recurrence. \Cref{fig:timeline} summarizes this ordering.

\paragraph{State handling and loss.}
States and zero-filled histories are reset at epoch and evaluation boundaries and whenever the batch size changes. This prevents state from one contiguous stream from leaking into an unrelated stream. The production loop detaches state, memory, and history tensors after every optimizer step. With one optimizer step per streaming batch, the effective truncated-backpropagation span is one batch transition \parencite{williams1990efficient}. A longer gradient accumulation appears in the demonstration script but is not part of production training. The production loss sums per-level BPS mean-squared error without level normalization. Consequently, hierarchies with different numbers of levels do not share a common loss scale, and a level with larger BPS variance may dominate the gradient:
\begin{equation}
 \mathcal{L}=\sum_{k=1}^{L}\frac{1}{B}\sum_{b=1}^{B}(\hat y_{b,t+1}^k-y_{b,t+1}^k)^2.
 \label{eq:loss}
\end{equation}
\texttt{StreamingBatcher} divides each chronological split into non-overlapping contiguous streams, discards any remainder that cannot fill the configured partition, and preserves chronological continuity at each batch position. Consecutive batches at a given stream position are therefore consecutive events in time, which is the continuity assumed by the masking and detachment behavior.

\paragraph{Training configuration.}
The production parser's defaults are batch size 256, eight evaluation streams, 50 maximum epochs, AdamW with learning rate $10^{-3}$ and weight decay $10^{-5}$ \parencite{loshchilov2019decoupled}, gradient clipping at 1.0 \parencite{pascanu2013difficulty}, patience 15, seed 42, cosine annealing with $T_{\max}=\text{epochs}$ \parencite{loshchilov2017sgdr}, and CUDA automatic mixed precision when available \parencite{micikevicius2018mixed}. Its source sets NumPy and PyTorch seeds but does not provide the ablation trainer's full Python-random and deterministic-cuDNN controls. The configuration table accompanying the reported results specifies the same batch size and epoch budget and adds the seeds 42, 151, and 359, whereas the separate ablation trainer defaults to batch size 256 and five epochs. These artifact-level configurations are inventoried in \Cref{tab:configinventory}, and \Cref{sec:provenance} states which is associated with each reported result. Validation inside the production trainer is stateful over its own contiguous streams, whereas the standalone evaluator uses a single stream and does not reproduce that default exactly.

\FloatBarrier

\section{Experimental Setup and Provenance}
\label{sec:protocol}
This section describes the data and comparators. Because several configurations coexist in the available artifacts, it also identifies the configuration associated with each reported result and states the caveat that governs the baseline comparison in later sections.

Per-seed aggregate metric files, the dataset configuration, the complete preprocessing pipeline, the completed-bar alignment and cleaning code, and seed-specific checkpoints are available in the project repository (\Cref{sec:availability}). Raw vendor market data cannot be redistributed, but the preprocessing pipeline is vendor-neutral and can regenerate the dataset from a compatible feed. Per-event predictions and persistence outputs are not included as precomputed files in the current release, although they can be generated from the released checkpoints and preprocessing pipeline. The zero-BPS persistence baseline remains a required rerun, and no values for it are reported here.

\subsection{Data, splits, and comparators}
The dataset covers four assets with different anchor frequencies and hierarchy depths: AAPL (M15/H1), EURUSD (M15/H1/H4), USDCHF (M15/M30/H1), and XAUUSD (M5/M15). This selection spans equities, foreign-exchange pairs, and gold, but it does not separate asset-specific effects from hierarchy-depth or period effects. Windows contain 32 bars by default unless explicitly overridden. Each sample is aligned to an anchor event and carries the latest completed window for every configured level, together with update indicators and next-bar targets, as formalized in \Cref{sec:problem}. \Cref{tab:dataset} reports sample counts, periods, and mean update rates.

\begin{table}[!htbp]\centering\scriptsize
\caption{Dataset and split statistics. Each row gives, for one asset and chronological split, the hierarchy, anchor timeframe, number of anchor-event observations, period, span in days, and mean update rate (the event-update rate averaged across levels). Per-level update rates are listed beneath the table.}\label{tab:dataset}
\begin{tabular}{llp{1.8cm}lrrrr}
\toprule Asset & Split & Hierarchy & Anchor & Observations & Period & Span (days) & Mean update\\\midrule
AAPL&Train&M15,H1&M15&43,331&2017-02-07 $\to$ 2023-09-27&2,423.1&0.616\\
AAPL&Val&M15,H1&M15&9,285&2023-09-27 $\to$ 2025-03-06&525.8&0.615\\
AAPL&Test&M15,H1&M15&9,286&2025-03-06 $\to$ 2026-08-10&522.2&0.615\\
EURUSD&Train&M15,H1,H4&M15&69,967&2022-07-26 $\to$ 2025-05-15&1,024.2&0.438\\
EURUSD&Val&M15,H1,H4&M15&14,992&2025-05-15 $\to$ 2025-12-19&218.2&0.438\\
EURUSD&Test&M15,H1,H4&M15&14,994&2025-12-19 $\to$ 2026-07-29&221.8&0.437\\
USDCHF&Train&M15,M30,H1&M15&69,975&2022-08-05 $\to$ 2025-05-28&1,026.8&0.583\\
USDCHF&Val&M15,M30,H1&M15&14,994&2025-05-28 $\to$ 2026-01-04&221.8&0.583\\
USDCHF&Test&M15,M30,H1&M15&14,996&2026-01-04 $\to$ 2026-08-11&218.2&0.583\\
XAUUSD&Train&M5,M15&M5&69,976&2025-03-13 $\to$ 2026-03-10&361.6&0.667\\
XAUUSD&Val&M5,M15&M5&14,994&2026-03-10 $\to$ 2026-05-26&77.4&0.667\\
XAUUSD&Test&M5,M15&M5&14,996&2026-05-26 $\to$ 2026-08-11&76.7&0.667\\
\bottomrule
\end{tabular}
\par\smallskip\parbox{0.95\linewidth}{\scriptsize Per-level update rates (finest to coarsest): AAPL M15/H1 = 1.00/0.232; EURUSD M15/H1/H4 = 1.00/0.250/0.063; USDCHF M15/M30/H1 = 1.00/0.500/0.250; XAUUSD M5/M15 = 1.00/0.333.}
\end{table}

The per-level update rates in \Cref{tab:dataset} show the expected step pattern: exactly 1.0 at the anchor level and decreasing at each coarser level, roughly in line with the ratio of bar durations. For example, USDCHF's H1 level closes once per four completed M15 bars, giving a rate of 0.250. This is a sanity check on the completed-bar alignment of \Cref{sec:problem}, consistent with the update indicator being computed correctly upstream of the masked transition; it is not an independent empirical claim. The counts, dates, and update rates are transcribed from the dataset configuration. The expected input format and cleaning procedure are described in \Cref{sec:availability}.

PatchTST uses $d_{\rm model}=128$, $d_{ff}=256$, two layers, eight heads, patch length 16, stride 8, disabled RevIN and decomposition, and horizon one \parencite{nie2023patchtst}. TimeXer shares these principal dimensions, patches the close price, and treats the remaining OHLCV channels as exogenous variables \parencite{wang2024timexer}. Both are single-timeframe architectures. They do not natively support multi-timeframe, event-driven state, so they are trained on the anchor timeframe, which is their native configuration. This places the two baselines within the single-timeframe architectural comparisons used in controlled multi-asset forecasting benchmarks \parencite{Saidd2026Benchmark}, whereas HARN is a multi-timeframe, event-driven model and is evaluated with all configured payload levels.

The comparison in \Cref{sec:results} is therefore a \emph{descriptive, scope-specific comparison of complete pipelines}. As \Cref{tab:fairness} shows, HARN receives every configured payload level and trains with stationary BPS targets, stateful contiguous streams, cosine scheduling, patience 15, and CUDA AMP when available. The baselines train on normalized absolute-close targets with constant learning rates, patience 5, stateless shuffled batches, and no AMP. This target difference is substantive: by \cref{eq:reconerr}, HARN's price error is a price-scaled BPS error anchored at the last observed close, whereas the baselines regress a normalized absolute close. The reported gap therefore does not isolate hierarchy, memory, resonance, target representation, training protocol, tuning effort, or architecture. This is the governing caveat for the baseline comparison.

\begin{table}[!htbp]\centering\scriptsize
\caption{Scope and training differences between HARN and the single-timeframe baseline trainers. PatchTST and TimeXer are trained on the anchor timeframe, their native configuration, whereas HARN is evaluated as a multi-timeframe model. These differences make the comparison in \Cref{tab:mainresults} descriptive: it does not isolate architecture, target definition, or training recipe.}\label{tab:fairness}
\begin{tabular}{>{\raggedright\arraybackslash}p{2.3cm}>{\raggedright\arraybackslash}p{3.5cm}>{\raggedright\arraybackslash}p{3.5cm}>{\raggedright\arraybackslash}p{3.5cm}}
\toprule Dimension&HARN&PatchTST&TimeXer\\\midrule
Scope&Multi-timeframe, event-driven&Single-timeframe, anchor-only&Single-timeframe, anchor-only\\
Input&5 OHLCV, all payload levels&5 OHLCV, anchor only&5 OHLCV, anchor only\\
Lookback / horizon&32 per level / 1&32 / 1&32 / 1\\
Training target&Stationary BPS delta&Normalized absolute-close target&Normalized absolute-close target\\
Batch/protocol&256 (reported recipe); stateful contiguous streams&256 shuffled; stateless&256 shuffled; stateless\\
Schedule; patience&Cosine; 15&Constant; 5&Constant; 5\\
AMP&CUDA AMP if available&No&No\\
Seed&Code default 42; aggregate results over 42, 151, 359&Configuration table records 42; per-seed CSVs cover 42, 151, 359&Configuration table records 42; per-seed CSVs cover 42, 151, 359\\
Selection&Best validation loss&Best validation loss&Best validation loss\\
\bottomrule
\end{tabular}
\end{table}

\FloatBarrier

\subsection{Which configuration produced which reported result}
\label{sec:provenance}
Because the available artifacts contain several configurations (\Cref{tab:configinventory}), \Cref{tab:provenance} records, for each reported item, the configuration that the manuscript's own evidence associates with it and what cannot be determined from the current materials. Where the materials are insufficient, the ambiguity is stated rather than resolved by inference.

\begin{table}[!htbp]\centering\scriptsize
\caption{Provenance of each reported item, based on statements in this manuscript and the repository artifacts. For each item, the table gives the associated configuration and what is and is not determined from the current materials.}\label{tab:provenance}
\begin{tabular}{>{\raggedright\arraybackslash}p{2.5cm}>{\raggedright\arraybackslash}p{4.9cm}>{\raggedright\arraybackslash}p{7.3cm}}
\toprule Reported item & Associated configuration & Determined / not determined\\\midrule
Dataset table (\Cref{tab:dataset}) & Dataset configuration; \texttt{dataset\_config} is available in the repository. & Values are transcribed. Raw vendor data cannot be redistributed, but the dataset can be regenerated from a compatible vendor feed with the vendor-neutral preprocessing pipeline.\\\addlinespace
HARN rows, Panels A and B (mean $\pm$ SD over three seeds) & Reported recipe: batch size 256, 50 epochs, AdamW ($10^{-3}$, $10^{-5}$), clipping 1.0, cosine schedule, patience 15, CUDA AMP if available; seeds 42, 151, 359. & The recipe coincides with the production defaults except for the seed list (\Cref{tab:configinventory}). Seed-specific checkpoints and per-seed aggregate metric CSVs are available, and per-event predictions can be generated from them with the preprocessing pipeline. The standalone evaluator uses one stream, whereas the recipe describes eight; the stream count behind the metrics should be read from the available configuration and checkpoint metadata. Whether Panels A and B came from the same checkpoints is documented in repository metadata where available and is otherwise unresolved.\\\addlinespace
PatchTST and TimeXer rows, Panel A & Baseline configuration of \Cref{tab:fairness}: single-timeframe, anchor-only, dimensions and patching as in \Cref{sec:protocol}, batch 256 shuffled, constant learning rate, patience 5, no AMP. & Per-seed aggregate metrics are available in the repository, and per-event predictions can be generated from the available pipeline and checkpoints. In Panel A, HARN and the baselines are reported at the anchor timeframe of each asset (M15 for AAPL, EURUSD, and USDCHF; M5 for XAUUSD).\\\addlinespace
Ablation variants A0--A3 (\Cref{tab:ablation}) & Ablation trainer defaults: batch 256, five epochs, seed 42, cosine schedule, patience 15, with configuration and seed retained in its checkpoint format. & The available evidence does not establish that these defaults were used for the table or identify repeated runs. A0 is full HARN within the ablation pipeline, but its values differ from the main HARN means and are not verified replicates; the reason is undetermined (\Cref{sec:ablation}).\\\addlinespace
\Cref{fig:representative,fig:errors} & Evaluation visualizations from the available outputs. & The captions do not identify asset, timeframe, seed, checkpoint, or split; the underlying predictions can be regenerated from the available checkpoints and preprocessing pipeline.\\\addlinespace
Audit table (\Cref{tab:audit}) & Static inspection of the source code. & Not an experiment (\Cref{sec:audit}).\\
\bottomrule
\end{tabular}
\end{table}

\FloatBarrier

\section{Metrics and Reported Results}
\label{sec:results}
For price errors $e_i=\hat P_i-P_i$, the evaluators report four point-forecast accuracy metrics that give both scale-dependent and scale-normalized views, consistent with the recommendation that no single measure suffices across series of different scales \parencite{hyndman2006another}:
\begin{align}
 \mathrm{MAE}&=n^{-1}\sum_i|e_i|, &
 \mathrm{RMSE}&=\sqrt{n^{-1}\sum_i e_i^2}, \\
 \mathrm{sMAPE}&=100\,n^{-1}\sum_i\frac{2|e_i|}{|P_i|+|\hat P_i|}, &
 \mathrm{MASE}&=\frac{\mathrm{MAE}}{\operatorname{mean}_j|y^{\rm train}_{j+1}-y^{\rm train}_j|}.
\end{align}
MAE and RMSE are in the asset's native price units, and sMAPE is a symmetric percentage error. The evaluator's MASE divides the reconstructed-price MAE by the mean absolute first difference of the stored training target-event sequence. Because higher-level target values can repeat while a coarser bar is carried forward, this denominator is an implementation-defined scale rather than necessarily the chronological close-series denominator of the standard MASE definition \parencite{hyndman2006another}. The evaluators do not emit a target-space metric, so all reported values are computed on reconstructed absolute price. The natural persistence comparator predicts $\hat y_i=0$ and therefore $\hat P_i=P_i^{\rm last}$ at each forecast origin. No persistence output is included as a precomputed file in the current release, and this comparator is a required rerun, not a missing value that can be inferred from the aggregate price metrics.

\Cref{tab:mainresults} reports the results. All values are transcribed as computed by the evaluators. The $\pm$ figures are aggregate variability across three seeds, whose per-seed outputs are given in \Cref{app:per-seed}, and \Cref{sec:provenance} records the configuration associated with each panel.

\begin{table}[!htbp]\centering\scriptsize
\caption{Forecasting results on the test split, reported as mean $\pm$ standard deviation across three seeds. Lower is better for every metric, and all metrics are computed on reconstructed absolute price. Panel A is a descriptive, scope-specific comparison between HARN (multi-timeframe, event-driven, BPS target, stateful training) and single-timeframe PatchTST and TimeXer trained on the anchor timeframe; it does not isolate architecture, target definition, or training recipe. Panel B reports HARN at further timeframes, for which no baseline was run; its XAUUSD M5 row repeats the Panel A entry. Values are reported as computed by the evaluators.}\label{tab:mainresults}
\setlength{\tabcolsep}{3.5pt}
\begin{tabular}{llrrrr}
\toprule Panel/asset & Model/TF & MAE & RMSE & sMAPE & MASE\\\midrule
\multicolumn{6}{l}{\textit{Panel A: HARN (multi-timeframe) versus single-timeframe anchor-only baselines (descriptive)}}\\
AAPL&HARN/M15&0.507696$\pm$0.000509&0.933489$\pm$0.000420&0.204947$\pm$0.000209&1.327584$\pm$0.001331\\
&PatchTST/M15&2.074401$\pm$0.575730&2.702157$\pm$0.559581&0.794647$\pm$0.219045&5.424392$\pm$1.505489\\
&TimeXer/M15&2.881886$\pm$0.811630&3.613453$\pm$0.848218&1.115299$\pm$0.292503&7.535901$\pm$2.122348\\
EURUSD&HARN/M15&0.000306$\pm$0.000002&0.000467$\pm$0.000002&0.026311$\pm$0.000188&0.865688$\pm$0.006210\\
&PatchTST/M15&0.000759$\pm$0.000275&0.000967$\pm$0.000305&0.065011$\pm$0.023455&2.144210$\pm$0.777650\\
&TimeXer/M15&0.010838$\pm$0.000347&0.013051$\pm$0.000521&0.930031$\pm$0.029837&30.626894$\pm$0.979317\\
USDCHF&HARN/M15&0.000263$\pm$0.000000&0.000391$\pm$0.000000&0.033284$\pm$0.000014&0.800131$\pm$0.000345\\
&PatchTST/M15&0.000638$\pm$0.000100&0.000864$\pm$0.000106&0.080930$\pm$0.012893&1.941151$\pm$0.304746\\
&TimeXer/M15&0.014144$\pm$0.000505&0.015651$\pm$0.000639&1.782636$\pm$0.063131&43.035375$\pm$1.536711\\
XAUUSD&HARN/M5&2.552111$\pm$0.001505&3.844300$\pm$0.001443&0.061098$\pm$0.000036&1.136012$\pm$0.000670\\
&PatchTST/M5&9.563464$\pm$1.040070&12.678673$\pm$0.945826&0.228356$\pm$0.025547&4.256949$\pm$0.462962\\
&TimeXer/M5&15.297477$\pm$1.745931&19.094934$\pm$2.925531&0.361264$\pm$0.038120&6.809308$\pm$0.777160\\
\addlinespace
\multicolumn{6}{l}{\textit{Panel B: HARN forecasts at further timeframes (no baseline available)}}\\
AAPL&H1&1.059066$\pm$0.000930&1.800225$\pm$0.000686&0.426064$\pm$0.000297&4.817277$\pm$0.004229\\
EURUSD&H1&0.000616$\pm$0.000002&0.000906$\pm$0.000002&0.052931$\pm$0.000188&3.463016$\pm$0.012255\\
EURUSD&H4&0.001227$\pm$0.000002&0.001745$\pm$0.000001&0.105431$\pm$0.000127&13.543420$\pm$0.016580\\
USDCHF&M30&0.000373$\pm$0.000001&0.000543$\pm$0.000000&0.047208$\pm$0.000065&1.607638$\pm$0.002219\\
USDCHF&H1&0.000526$\pm$0.000001&0.000763$\pm$0.000000&0.066632$\pm$0.000122&3.232735$\pm$0.005914\\
XAUUSD&M5&2.552111$\pm$0.001505&3.844300$\pm$0.001443&0.061098$\pm$0.000036&1.136012$\pm$0.000670\\
\bottomrule
\end{tabular}
\setlength{\tabcolsep}{4pt}
\end{table}

\subsection{Baseline comparison and further timeframes}
In Panel A of \Cref{tab:mainresults}, HARN's price errors are numerically lower than those of both baselines for all four assets at the anchor timeframe: M15 for AAPL, EURUSD, and USDCHF, and M5 for XAUUSD. The XAUUSD HARN entries equal the means of the per-seed M5 values in \Cref{tab:harn-per-seed}. As \Cref{sec:protocol} explains, this is a statement about the configured pipelines (\Cref{tab:fairness}), not an architectural result. No reported run varies the available timeframes, target representation, or training protocol one factor at a time. The results therefore do not identify whether a numerical difference arises from additional timeframes, the BPS target, stateful streaming, tuning effort, or architecture. Three further points guide interpretation.

\paragraph{Missing persistence baseline.}
The reported results contain no persistence (last-value, that is, zero-BPS) or other random-walk baseline. Because HARN's price error is a price-scaled BPS error anchored at the last close (\cref{eq:reconerr}), such a reference is the natural yardstick, and its absence leaves the results without an interpretability baseline. The MASE column does not fill this role. Its denominator is an implementation-defined training-sample scale, and HARN's own MASE values range from below one (EURUSD and USDCHF at the anchor) to above one (the AAPL and XAUUSD anchor rows and all higher levels), so they do not settle whether HARN improves on persistence over the test period.

\paragraph{Omitted percentage reductions.}
The original results also expressed HARN's advantage over PatchTST as percentage reductions, which are omitted here. They restate Panel A as an effect-size-like quantity for a scope-specific comparison and so invite an attribution to architecture that the design cannot support. They are also ratios of aggregate means without uncertainty, their aggregation method is undocumented, and a few entries differ by about 0.1 percentage points from the reduction implied by the rounded tabulated means. Nothing has been recomputed or replaced.

\paragraph{Further timeframes.}
Panel B reports HARN's forecasts at the further levels of each hierarchy, which have different update rates (\Cref{tab:dataset}). It documents forecasts at several resolutions and does not show superiority, cross-level forecast coherence, or benefit from the hierarchy. Its XAUUSD M5 row repeats the anchor entry of Panel A, and per-seed results for the non-anchor XAUUSD M15 level appear in \Cref{tab:harn-per-seed}. MASE denominators at higher levels use repeated targets, so MASE values are not comparable across timeframes.

\subsection{Cross-seed variance and statistical status}
The reported HARN standard deviations across seeds are much smaller than those of the baselines. For AAPL MAE, HARN's standard deviation is about 0.1\% of its mean, compared with about 28\% for both PatchTST and TimeXer; for EURUSD MAE, the corresponding figures are below 1\% for HARN and about 36\% for PatchTST. Several HARN entries round to $\pm0.000000$ at the reported precision, including USDCHF MAE and RMSE in Panel A and RMSE at M30 and H1 in Panel B. The pattern appears across metrics and panels.

We regard this as an unresolved variance observation: the available evidence does not establish its cause, and we do not offer one. Seed-specific checkpoints and per-seed aggregate metrics are available (\Cref{app:per-seed}), but per-seed training trajectories and per-event predictions are not included as precomputed files in the current release. Seed independence therefore cannot be verified from the CSVs and checkpoints alone; demonstrating that the three seeds (42, 151, 359) produced independent training trajectories requires inspecting the seed-specific checkpoints and their configuration metadata. The small HARN spread should not be read as evidence of robustness, and the baselines' larger spread should not be read as a property of those architectures, given the differing protocols. Interpreting either requires the per-event outputs and the verification identified here as necessary follow-up work.

\begin{figure}[!htbp]
\centering
\includegraphics[width=0.95\linewidth]{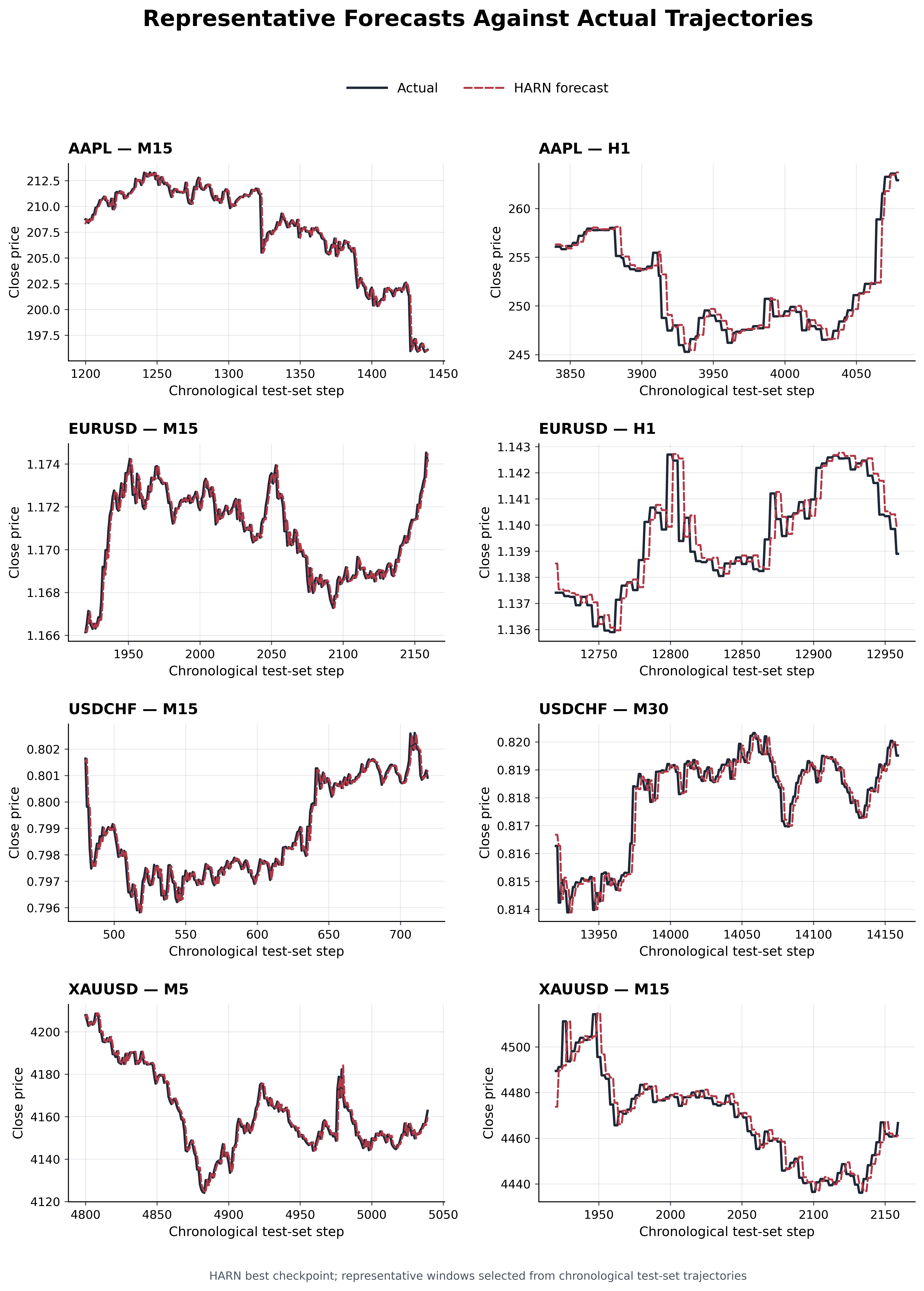}
\caption{Representative forecasts against actual trajectories, taken from the evaluation outputs. The asset, timeframe, seed, and split are not identified in the available materials, so the figure gives qualitative context only.}
\label{fig:representative}
\end{figure}

\begin{figure}[!htbp]
\centering
\includegraphics[width=0.95\linewidth]{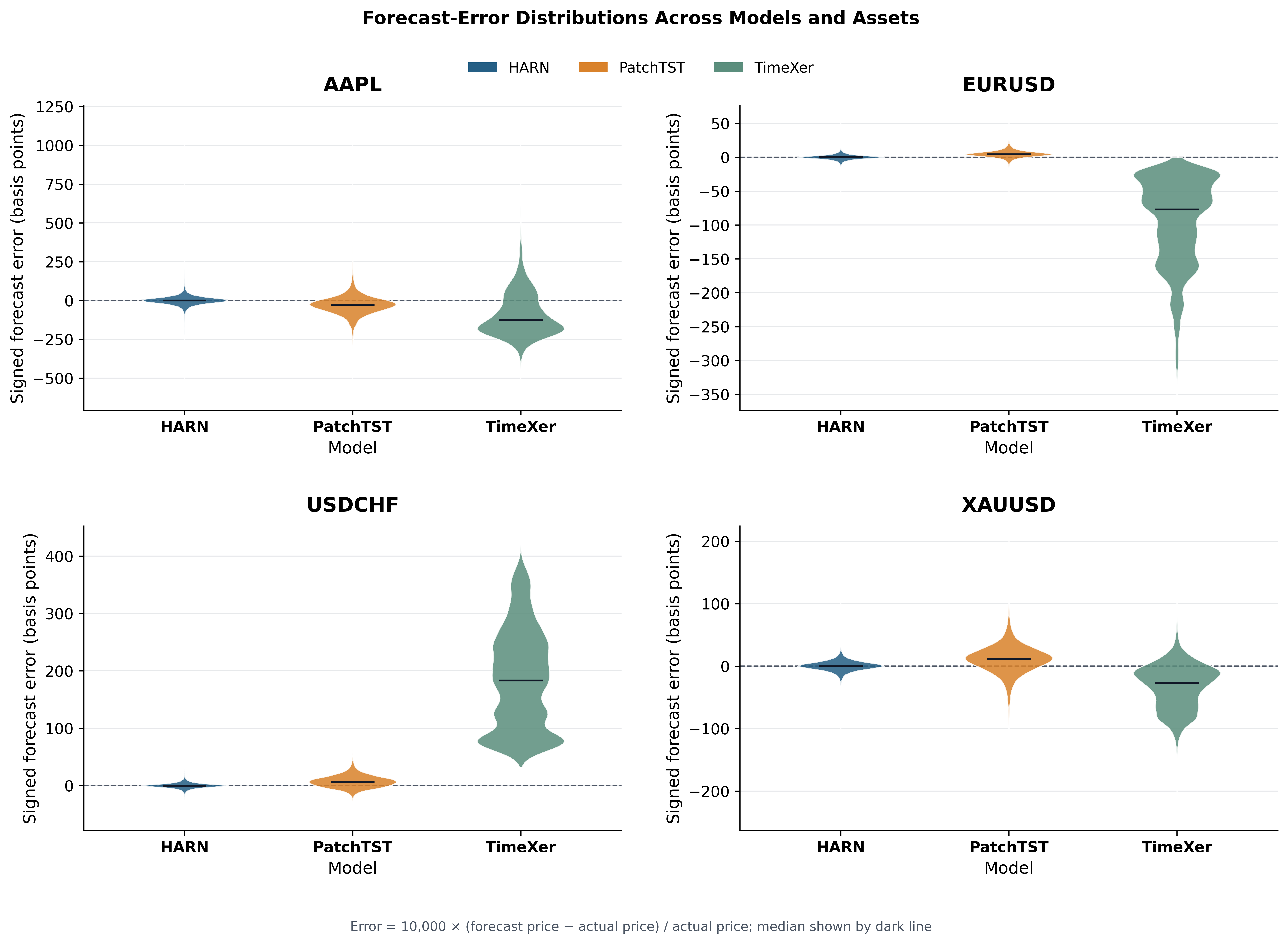}
\caption{Representative forecast-error distributions across models and assets, taken from the evaluation outputs. The provenance limits are the same as for \Cref{fig:representative}, and the figure gives qualitative context only.}
\label{fig:errors}
\end{figure}

\Cref{fig:representative,fig:errors} are empirical visualizations taken from the evaluation outputs, in contrast with the conceptual architecture and timeline figures. Their captions do not identify the asset, timeframe, seed, or split shown, and the underlying predictions are not included as precomputed files in the current release, so the figures provide qualitative context only and no statistical comparison.

\FloatBarrier

A standard deviation over three aggregate seed scores is not a confidence interval for the temporal forecast-error process, does not capture dependence across forecast origins, and cannot recover the paired model differences required by a formal predictive-accuracy test \parencite{diebold1995comparing,giacomini2006tests}. No confidence intervals, paired tests, Diebold--Mariano tests, Wilcoxon tests, effect sizes, or multiplicity corrections were computed, and none are introduced here. \Cref{sec:discussion} identifies these analyses as proposed work.

\section{Ablation Variants: Single-Run Observations}
\label{sec:ablation}
The ablation study asks a narrower question than the baseline comparison: how do HARN's own reported values change when one named internal mechanism is replaced or bypassed, with the surrounding streaming pipeline and tensor interfaces held fixed? The variants share HARN's interfaces and pipeline but not its parameterization or parameter count, so they are component-replacement interventions rather than matched counterfactuals. A0 is full HARN. A1 replaces the associative cell of \Cref{sec:memory} with a gated recurrent-style update while carrying the memory tensor unchanged. A2 replaces resonance (\cref{eq:resonance}) with the identity map while retaining bottom-up evidence, so it removes only the all-to-all path and not all cross-level communication. A3 replaces the history readout (\cref{eq:evidence}) with zero evidence while preserving the interface and history maintenance. Each intervention can alter capacity and optimization dynamics in addition to the named mechanism: A1 changes the cell's parameterization, A2 removes learnable resonance parameters, and A3 removes learned evidence content.

\Cref{tab:ablation} covers only AAPL and EURUSD and is rounded to six decimals. Each cell is a single aggregate value with no uncertainty estimate, and the available evidence does not document repeated runs, so the entries are treated as single-run observations rather than statistically established sensitivity effects. For metric $m$, the relative degradation is defined as
\begin{equation}
 \Delta_i=\frac{m(A_i)-m(A_0)}{|m(A_0)|},
 \label{eq:degradation}
\end{equation}
where a positive value indicates higher error than the full model.

\begin{table}[!htbp]\centering\scriptsize
\caption{Ablation variants for AAPL and EURUSD only. Each cell is a single aggregate value without per-seed uncertainty, and lower is better. A0 is full HARN; A1 replaces the associative memory with a gated recurrent update; A2 replaces resonance with the identity map; A3 zeros the bottom-up evidence readout. Configuration provenance is given in \Cref{tab:provenance}.}\label{tab:ablation}
\begin{tabular}{lllrrrr}
\toprule Asset&TF&Variant&MAE&RMSE&sMAPE&MASE\\\midrule
AAPL&H1&A0&1.055134&1.801484&0.424678&4.799395\\
AAPL&H1&A1&1.056989&1.798886&0.425441&4.807829\\
AAPL&H1&A2&1.060339&1.803577&0.426672&4.823069\\
AAPL&H1&A3&1.058882&1.796596&0.426056&4.816440\\
\addlinespace
AAPL&M15&A0&0.506070&0.932352&0.204299&1.326365\\
AAPL&M15&A1&0.507230&0.933322&0.204760&1.323333\\
AAPL&M15&A2&0.507286&0.933827&0.204801&1.326513\\
AAPL&M15&A3&0.508024&0.934081&0.205084&1.328443\\
\addlinespace
EURUSD&H1&A0&0.000610&0.000902&0.052438&3.469651\\
EURUSD&H1&A1&0.000617&0.000907&0.053033&3.430735\\
EURUSD&H1&A2&0.000614&0.000903&0.052699&3.447709\\
EURUSD&H1&A3&0.000613&0.000904&0.052646&3.444383\\
\addlinespace
EURUSD&H4&A0&0.001226&0.001742&0.105436&13.534480\\
EURUSD&H4&A1&0.001231&0.001747&0.105728&13.581712\\
EURUSD&H4&A2&0.001229&0.001746&0.105552&13.559399\\
EURUSD&H4&A3&0.001227&0.001743&0.105364&13.544075\\
\addlinespace
EURUSD&M15&A0&0.000305&0.000465&0.026176&0.861235\\
EURUSD&M15&A1&0.000306&0.000466&0.026240&0.863367\\
EURUSD&M15&A2&0.000305&0.000465&0.026183&0.861477\\
EURUSD&M15&A3&0.000306&0.000466&0.026250&0.863671\\
\bottomrule
\end{tabular}
\end{table}

\begin{figure}[!htbp]
\centering
\includegraphics[width=0.96\linewidth]{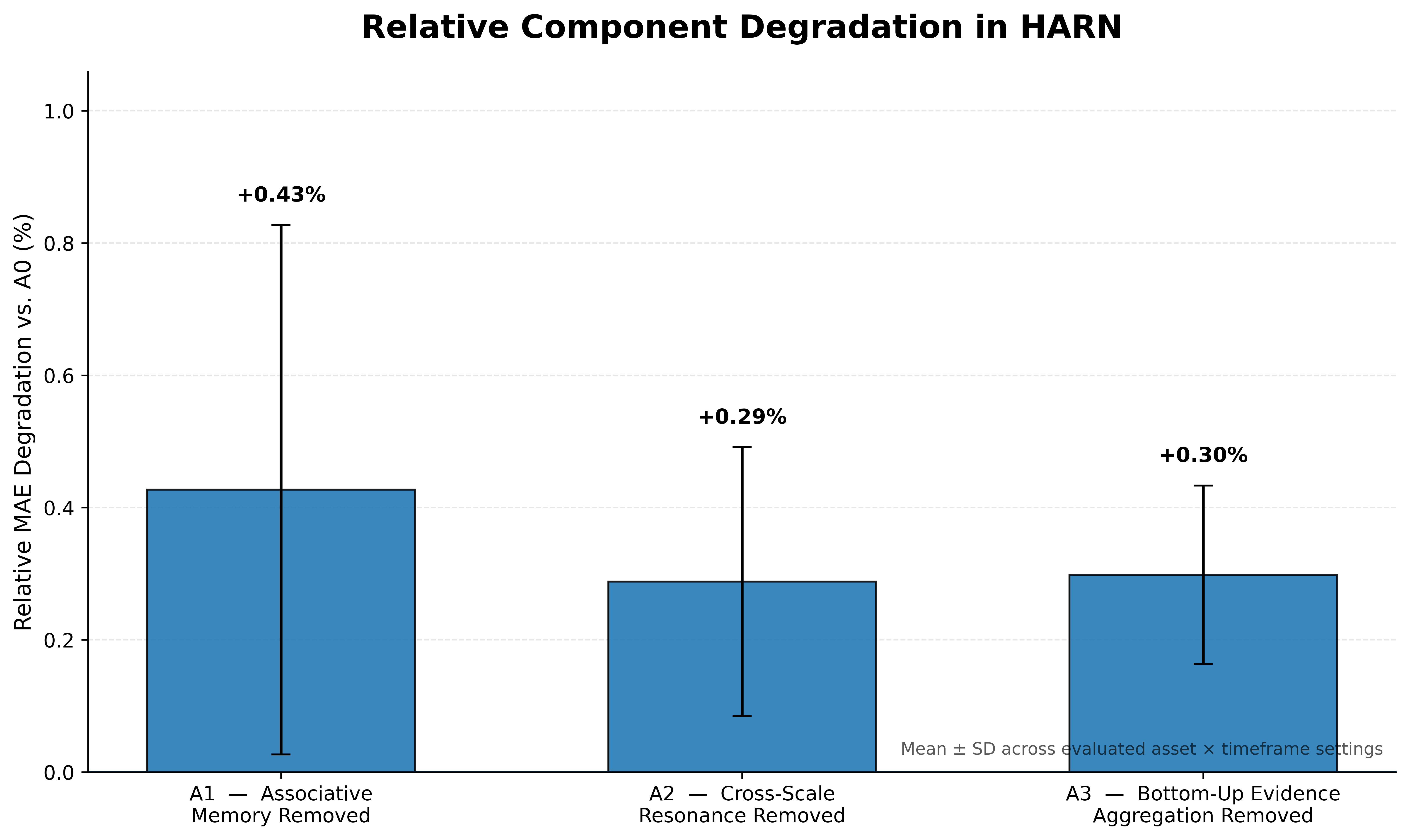}
\caption{Relative component degradation $\Delta_i$ (\cref{eq:degradation}) for each ablation variant, asset, and timeframe, computed from \Cref{tab:ablation}. The plot summarizes single aggregate values; it is descriptive and carries no uncertainty estimate.}
\label{fig:degradation}
\end{figure}

Two features of \Cref{tab:ablation} limit what it can show. First, the differences are small and non-monotonic. A0 is not uniformly best: A1 and A3 have lower AAPL H1 RMSE than A0, and A1--A3 all have lower EURUSD H1 MASE. A variant scoring below A0 on one metric does not by itself indicate that the removed mechanism is harmful, given the capacity and optimization differences described above. Second, A0 does not coincide with the corresponding HARN means in \Cref{tab:mainresults}. For AAPL M15, the A0 MAE is 0.506070 against a Panel A mean of 0.507696, and for AAPL H1 it is 1.055134 against a Panel B mean of 1.059066. These gaps are of similar size to the gaps between A0 and A1--A3, and the manuscript does not establish whether A0 is a rerun, a different epoch budget, or a different evaluation path. Variant differences of this size therefore cannot currently be separated from unexplained run-to-run or protocol differences. A measured reading is that no single component dominates these single-run observations, and that firmer statements would require the multi-seed, capacity-matched experiments identified as proposed work. \Cref{fig:degradation} plots $\Delta_i$ and is descriptive only.

\FloatBarrier

\section{Code-Level Causality Audit}
\label{sec:audit}
This section records a static review of the source code against the event-level causal contract defined in \Cref{sec:problem}. It is a code-level causal-consistency audit and \emph{not} an empirical test: no experiment was conducted in which bars after $\tau$ were perturbed, shuffled, or withheld to verify that forecasts remain unchanged. Code review also cannot detect upstream problems in the data itself, such as source-level bar construction, timestamp conventions, clock synchronization, or price adjustment. An empirical check of this kind remains proposed work. \Cref{tab:audit} summarizes the audit of chronological splitting, completed-bar alignment, next-bar targets, training-only statistics, left padding, shifted derived features, state masking, reset and detachment behavior, contiguous streams, cleaning, and validation-based checkpoint selection, following the implementation-level considerations relevant to leakage and validation in the forecasting literature \parencite{kaufman2012leakage,tashman2000out,bergmeir2018validity}. The findings indicate consistency with the specified future-event information-ordering contract under the implementation's event ordering. Reproducibility is supported by the artifacts listed in \Cref{sec:availability}, subject to the raw-data redistribution constraint.

Three qualifications apply. First, the lower-level state is appended before the higher-level evidence is read within the same event. The evidence path is thus ordered with respect to future events but is not strictly past-event-only within an event; this is the same-event evidence-ordering qualification of \Cref{sec:problem}, recorded in row~9 of \Cref{tab:audit}. Second, inactive levels are state-preserving but not computation-free: their windows and anchors are processed at every event, and only the persistent overwrite is masked (\cref{eq:maskedtransition}). Third, the exact-timestamp-equality rule underlying $u_\tau^k$ is a data-interface assumption. Feeds with different timestamp conventions, clock skew, or asynchronous bar-close reporting would require an explicit alignment policy before the same audit could be applied unchanged.

{\small
\setlength{\tabcolsep}{3pt}
\begin{longtable}{%
    >{\raggedright\arraybackslash}p{0.5cm}
    >{\raggedright\arraybackslash}p{2.3cm}
    >{\raggedright\arraybackslash}p{7.3cm}
    >{\raggedright\arraybackslash}p{2.3cm}
}
\caption{Condensed code-level causal-consistency audit from a static review of the source code; it is not an empirical causality test. ``Consistent (qualified)'' denotes consistency with the event-level information-ordering contract, subject to the same-event evidence-ordering qualification of \Cref{sec:problem}.}
\label{tab:audit}\\
\toprule
\# & Dimension & Mechanism observed in the code & Finding \\
\midrule
\endfirsthead

\toprule
\# & Dimension & Mechanism observed in the code & Finding \\
\midrule
\endhead

1 & Splitting
& Chronological 70/15/15 train--validation--test slices
& Consistent \\

2 & Alignment
& Latest completed timestamp no later than the anchor; exact timestamp equality determines an update
& Consistent \\

3 & Windows/\newline targets
& Window ends at the selected completed bar; target is the next completed bar
& Consistent \\

4 & Statistics
& Training-only feature and target statistics are reused during evaluation
& Consistent \\

5 & Local causality
& Left-only padding and shifted previous-close features avoid dependence on future observations
& Consistent \\

6 & Resonance
& Only states available at the same event are mixed
& Consistent \\

7 & Memory
& Current input, previous state, and bounded residual-driven write determine memory updates
& Consistent \\

8 & Masked updates
& Inactive higher-level persistent states and memory are retained rather than overwritten
& Consistent \\

9 & History
& Current lower-level read is appended before higher-level readout
& Consistent (qualified) \\

10 & Lifecycle
& States are reset at boundaries and states and histories are detached after steps
& Consistent \\

11 & Streaming
& Non-overlapping chronological streams preserve batch-position continuity
& Consistent \\

12 & Cleaning
& Invalid rows are removed, duplicates are resolved, and timestamps are sorted chronologically
& Consistent \\

13 & Selection
& Validation loss selects the checkpoint before test evaluation
& Consistent \\

14 & Test isolation
& Test observations are not used for training or validation
& Consistent \\

\bottomrule
\end{longtable}
}

\section{Discussion and Limitations}
\label{sec:discussion}

\paragraph{What the evidence supports.}
HARN is an event-driven, multi-timeframe architecture that maintains a compact state for each configured resolution and updates that state when new information becomes available at the corresponding resolution. Its principal mechanisms are residual-driven associative memory, cross-scale resonance, and ordered history readout. Together, they retain, exchange, and read information across asynchronously completing timeframes. The experiments characterize this design under the specified completed-bar protocol, and the code-level audit finds the implementation consistent with the event-driven alignment rules used in the experiments. Under the reported configuration, HARN achieves lower reconstructed-price errors than the accompanying single-timeframe, anchor-only baselines on the evaluated datasets. That comparison remains descriptive, however, because the models differ in information access, target formulation, and training procedure. The ablations show only small changes when individual components are replaced, but their single-run design does not establish independent component effects. The results therefore support the viability of the design in the evaluated setting; they do not establish superiority over alternative architectures or attribute the observed differences to any one design choice.

\paragraph{Limitations.}
The main limitations concern comparison scope, reference baselines, ablation strength, generalization, computational evaluation, and design assumptions.
\begin{itemize}[leftmargin=*,itemsep=2pt]
\item \emph{Comparison scope.} The baseline comparison does not isolate architectural effects from differences in information access, target formulation, or training procedure, and it is therefore not a controlled architectural comparison.
\item \emph{Reference baselines and statistics.} No persistence baseline, confidence intervals, or paired predictive-accuracy tests are reported, and the unusually small cross-seed spread of HARN remains unexplained.
\item \emph{Ablation strength.} The ablations are single-run experiments on a subset of assets and are not parameter- or compute-matched. They provide component-level observations but not a statistically robust attribution of performance to individual mechanisms.
\item \emph{Generalization.} The study covers four assets and specific evaluation periods. The available evidence does not establish how performance changes across a broader range of assets, market conditions, hierarchy depths, or time periods.
\item \emph{Computational evaluation.} The study does not report systematic measurements of inference latency, peak memory, throughput, or energy consumption, so the practical cost of maintaining and updating multiple resolution-specific states remains incompletely characterized.
\item \emph{Design assumptions.} The evidence-readout coefficients are hand-set and unvalidated, and the update indicator relies on exact timestamp equality under the chosen bar-construction protocol.
\end{itemize}

\paragraph{Proposed experiments.}
The following studies have not been performed and constitute proposed work: (i) evaluation of the zero-BPS persistence baseline; (ii) generation of per-event predictions from the available checkpoints to enable confidence intervals and paired tests such as Diebold--Mariano and conditional predictive-ability tests; (iii) verification of seed independence from checkpoint metadata and training trajectories; (iv) controlled comparisons that vary the available timeframes, target representation, and training protocol one factor at a time, together with matched multi-timeframe or event-driven baselines; (v) multi-seed, capacity-matched ablations, including sensitivity to the hand-set evidence coefficients; (vi) an empirical perturbation test in which bars after the anchor time are altered or withheld; (vii) systematic measurement of latency, memory, and throughput; and (viii) archiving generated outputs with command lines, source hashes, data checksums, and software and device versions.

\section{Conclusion}
\label{sec:conclusion}
This paper specifies HARN as an event-driven, stateful multi-timeframe forecaster combining causal local encoding, gated matrix memory, all-to-all cross-scale resonance, directional bottom-up evidence, masked asynchronous updates, and BPS reconstruction. Its central contribution is the completed-bar update protocol, formalized in scalar and batched form and examined through a code-level event-causality audit. The reported results show numerically lower absolute-price error for HARN than for the accompanying PatchTST and TimeXer runs. Those baselines are single-timeframe models trained on the anchor timeframe in their native configuration, whereas HARN is evaluated as a multi-timeframe, event-driven model. The comparison is therefore descriptive and scope-specific; it does not isolate architecture, target definition, or training recipe. The single-run ablations identify no dominant component. Whether the reported differences reflect multi-timeframe inputs, target representation, training procedure, or their interaction remains open. Addressing this question requires the broader comparisons, persistence evaluation, per-event predictions, and paired statistical analyses described in \Cref{sec:discussion}.

\section{Data and Code Availability}
\label{sec:availability}
\paragraph{Repository artifacts.}
The project repository contains all source code examined in this study, including the HARN implementation; data-cleaning and completed-bar alignment code; the preprocessing pipeline; dataset configuration; the production training script; a standalone evaluator; an ablation trainer; and the PatchTST and TimeXer trainers. It also includes model checkpoints for each seed and per-seed result CSV files. The candidate public URL is \url{https://github.com/NabeelAhmadSaidd/HARN}.

\paragraph{Data sharing and vendor-agnostic acquisition.}
The raw market data used in the reported runs cannot be redistributed because the author does not have permission to share the vendor-provided data. The preprocessing pipeline is vendor-neutral and can be applied to compatible data from another market-data vendor. The expected input is one row per completed bar and timeframe, with columns for timestamp, open, high, low, close, and volume. Timestamps must be parseable and use a consistent timezone and session convention; bar construction must follow the configured per-level timeframes (for example, M5, M15, M30, H1, and H4). Cleaning parses timestamps, removes invalid rows, retains the last occurrence of duplicate timestamps, and sorts rows in ascending order. Completed-bar alignment selects the latest level-$k$ bar whose timestamp is no later than the anchor timestamp, and the update indicator is set by exact timestamp equality. Splits are chronological 70\%/15\%/15\%, and feature and target statistics are computed on training samples only. A compatible vendor feed with the same schema and bar-construction rules can therefore regenerate the dataset with the provided pipeline.

\paragraph{Checkpoints, evaluator, and generated outputs.}
The production trainer writes seed-specific checkpoints. The standalone evaluator reconstructs missing configuration fields from defaults and tensor shapes and evaluates on a single stream rather than the trainer's eight-stream default. Ablation checkpoints retain configuration metadata but are not parameter-count matched across variants. Per-event predictions and persistence outputs are not included as precomputed files in the current per-seed CSV release, but they can be generated from the available checkpoints and preprocessing pipeline with a compatible vendor feed. Archiving those generated outputs alongside the exact command line, source hash, data checksums, Python, PyTorch, and CUDA versions, and device details remains proposed work.

\paragraph{Command templates.}
The source-level command templates are as follows: production HARN training with \texttt{python -m src.models.train\_harn} and \texttt{--asset ASSET}; standalone evaluation with \texttt{python -m src.models.evaluate\_harn}, \texttt{--asset ASSET}, and \texttt{--seed SEED}; and ablation training with \texttt{python -m src.ablation.train\_harn\_ablations}, \texttt{--asset ASSET}, and \texttt{--variant A0} through \texttt{A3}. These templates are documented commands, not recovered historical commands for the reported tables.

\section*{Conflicts of Interest}
The author declares that there are no conflicts of interest associated with this work. The research was conducted independently, with no financial, commercial, institutional, or personal interests that could reasonably be considered to have influenced the design, implementation, evaluation, or interpretation of the study.

No external organization provided funding, directed the experimental methodology, or exercised control over the reported results. The work is presented solely for research purposes and does not constitute investment advice, a recommendation to trade any financial instrument, or an endorsement of any particular financial product or service.

\printbibliography

@article{fama1970efficient,
  author  = {E. F. Fama},
  title   = {Efficient Capital Markets: A Review of Theory and Empirical Work},
  journal = {The Journal of Finance},
  year    = {1970},
  volume  = {25},
  number  = {2},
  pages   = {383--417},
  doi     = {10.1111/j.1540-6261.1970.tb00518.x}
}

@article{cont2001empirical,
  author  = {R. Cont},
  title   = {Empirical Properties of Asset Returns: Stylized Facts and Statistical Issues},
  journal = {Quantitative Finance},
  year    = {2001},
  volume  = {1},
  number  = {2},
  pages   = {223--236},
  doi     = {10.1088/1469-7688/1/2/304}
}

@book{tsay2010analysis,
  author    = {R. S. Tsay},
  title     = {Analysis of Financial Time Series},
  edition   = {3},
  publisher = {John Wiley \& Sons},
  address   = {Hoboken, NJ},
  year      = {2010},
  isbn      = {9780470414354},
  doi       = {10.1002/9780470644560}
}

@book{hamilton1994time,
  author    = {J. D. Hamilton},
  title     = {Time Series Analysis},
  publisher = {Princeton University Press},
  address   = {Princeton, NJ},
  year      = {1994},
  isbn      = {9780691042893},
  doi       = {10.1515/9780691218632}
}

@article{engle1982arch,
  author  = {R. F. Engle},
  title   = {Autoregressive Conditional Heteroscedasticity with Estimates of the Variance of United Kingdom Inflation},
  journal = {Econometrica},
  year    = {1982},
  volume  = {50},
  number  = {4},
  pages   = {987--1007},
  doi     = {10.2307/1912773}
}

@article{bollerslev1986generalized,
  author  = {T. Bollerslev},
  title   = {Generalized Autoregressive Conditional Heteroskedasticity},
  journal = {Journal of Econometrics},
  year    = {1986},
  volume  = {31},
  number  = {3},
  pages   = {307--327},
  doi     = {10.1016/0304-4076(86)90063-1}
}

@article{kaufman2012leakage,
  author  = {S. Kaufman and S. Rosset and C. Perlich and O. Stitelman},
  title   = {Leakage in Data Mining: Formulation, Detection, and Avoidance},
  journal = {ACM Transactions on Knowledge Discovery from Data},
  year    = {2012},
  volume  = {6},
  number  = {4},
  pages   = {15:1--15:21},
  doi     = {10.1145/2382577.2382579}
}

@article{tashman2000out,
  author  = {L. J. Tashman},
  title   = {Out-of-Sample Tests of Forecasting Accuracy: An Analysis and Review},
  journal = {International Journal of Forecasting},
  year    = {2000},
  volume  = {16},
  number  = {4},
  pages   = {437--450},
  doi     = {10.1016/S0169-2070(00)00065-0}
}

@article{bergmeir2018validity,
  author  = {C. Bergmeir and R. J. Hyndman and B. Koo},
  title   = {A Note on the Validity of Cross-Validation for Evaluating Autoregressive Time Series Prediction},
  journal = {Computational Statistics \& Data Analysis},
  year    = {2018},
  volume  = {120},
  pages   = {70--83},
  doi     = {10.1016/j.csda.2017.11.003}
}

@article{hyndman2006another,
  author  = {R. J. Hyndman and A. B. Koehler},
  title   = {Another Look at Measures of Forecast Accuracy},
  journal = {International Journal of Forecasting},
  year    = {2006},
  volume  = {22},
  number  = {4},
  pages   = {679--688},
  doi     = {10.1016/j.ijforecast.2006.03.001}
}

@article{diebold1995comparing,
  author  = {F. X. Diebold and R. S. Mariano},
  title   = {Comparing Predictive Accuracy},
  journal = {Journal of Business \& Economic Statistics},
  year    = {1995},
  volume  = {13},
  number  = {3},
  pages   = {253--263},
  doi     = {10.1080/07350015.1995.10524599}
}

@article{giacomini2006tests,
  author  = {R. Giacomini and H. White},
  title   = {Tests of Conditional Predictive Ability},
  journal = {Econometrica},
  year    = {2006},
  volume  = {74},
  number  = {6},
  pages   = {1545--1578},
  doi     = {10.1111/j.1468-0262.2006.00718.x}
}

@article{gneiting2007strictly,
  author  = {T. Gneiting and A. E. Raftery},
  title   = {Strictly Proper Scoring Rules, Prediction, and Estimation},
  journal = {Journal of the American Statistical Association},
  year    = {2007},
  volume  = {102},
  number  = {477},
  pages   = {359--378},
  doi     = {10.1198/016214506000001437}
}

@inproceedings{vaswani2017attention,
  author    = {A. Vaswani and N. Shazeer and N. Parmar and J. Uszkoreit and L. Jones and A. N. Gomez and L. Kaiser and I. Polosukhin},
  title     = {Attention Is All You Need},
  booktitle = {Advances in Neural Information Processing Systems},
  volume    = {30},
  pages     = {5998--6008},
  publisher = {Curran Associates, Inc.},
  year      = {2017},
  doi       = {10.48550/arXiv.1706.03762}
}

@inproceedings{zhou2021informer,
  author    = {H. Zhou and S. Zhang and J. Peng and S. Zhang and J. Li and H. Xiong and W. Zhang},
  title     = {Informer: Beyond Efficient Transformer for Long Sequence Time-Series Forecasting},
  booktitle = {Proceedings of the AAAI Conference on Artificial Intelligence},
  volume    = {35},
  number    = {12},
  pages     = {11106--11115},
  publisher = {AAAI Press},
  year      = {2021},
  doi       = {10.1609/aaai.v35i12.17325}
}

@inproceedings{wu2021autoformer,
  author    = {H. Wu and J. Xu and J. Wang and M. Long},
  title     = {Autoformer: Decomposition Transformers with Auto-Correlation for Long-Term Series Forecasting},
  booktitle = {Advances in Neural Information Processing Systems},
  volume    = {34},
  pages     = {22419--22430},
  publisher = {Curran Associates, Inc.},
  year      = {2021},
  doi       = {10.48550/arXiv.2106.13008}
}

@inproceedings{zhou2022fedformer,
  author    = {T. Zhou and Z. Ma and Q. Wen and X. Wang and L. Sun and R. Jin},
  title     = {FEDformer: Frequency Enhanced Decomposed Transformer for Long-Term Series Forecasting},
  booktitle = {Proceedings of the 39th International Conference on Machine Learning (ICML)},
  series    = {Proceedings of Machine Learning Research},
  volume    = {162},
  pages     = {27268--27286},
  publisher = {PMLR},
  year      = {2022},
  doi       = {10.48550/arXiv.2201.12740}
}

@inproceedings{nie2023patchtst,
  author    = {Y. Nie and N. H. Nguyen and P. Sinthong and J. Kalagnanam},
  title     = {A Time Series is Worth 64 Words: Long-Term Forecasting with Transformers},
  booktitle = {The Eleventh International Conference on Learning Representations (ICLR)},
  publisher = {OpenReview.net},
  year      = {2023},
  doi       = {10.48550/arXiv.2211.14730}
}

@inproceedings{woo2023timesnet,
  author    = {H. Wu and T. Hu and Y. Liu and H. Zhou and J. Wang and M. Long},
  title     = {TimesNet: Temporal 2D-Variation Modeling for General Time Series Analysis},
  booktitle = {The Eleventh International Conference on Learning Representations (ICLR)},
  publisher = {OpenReview.net},
  year      = {2023},
  doi       = {10.48550/arXiv.2210.02186}
}

@inproceedings{wang2024timexer,
  author    = {Y. Wang and H. Wu and J. Dong and G. Qin and H. Zhang and Y. Liu and Y. Qiu and J. Wang and M. Long},
  title     = {TimeXer: Empowering Transformers for Time Series Forecasting with Exogenous Variables},
  booktitle = {Advances in Neural Information Processing Systems},
  volume    = {37},
  pages     = {469--498},
  publisher = {Curran Associates, Inc.},
  year      = {2024},
  doi       = {10.52202/079017-0015}
}

@article{hochreiter1997long,
  author  = {S. Hochreiter and J. Schmidhuber},
  title   = {Long Short-Term Memory},
  journal = {Neural Computation},
  year    = {1997},
  volume  = {9},
  number  = {8},
  pages   = {1735--1780},
  doi     = {10.1162/neco.1997.9.8.1735}
}

@inproceedings{voelker2019legendre,
  author    = {A. Voelker and I. Kaji\'c and C. Eliasmith},
  title     = {Legendre Memory Units: Continuous-Time Representation in Recurrent Neural Networks},
  booktitle = {Advances in Neural Information Processing Systems},
  volume    = {32},
  publisher = {Curran Associates, Inc.},
  year      = {2019}
}

@inproceedings{gu2022efficient,
  author    = {A. Gu and K. Goel and C. R\'e},
  title     = {Efficiently Modeling Long Sequences with Structured State Spaces},
  booktitle = {The Tenth International Conference on Learning Representations (ICLR)},
  publisher = {OpenReview.net},
  year      = {2022},
  doi       = {10.48550/arXiv.2111.00396}
}

@inproceedings{smith2023simplified,
  author    = {J. T. H. Smith and A. Warrington and S. W. Linderman},
  title     = {Simplified State Space Layers for Sequence Modeling},
  booktitle = {The Eleventh International Conference on Learning Representations (ICLR)},
  publisher = {OpenReview.net},
  year      = {2023},
  doi       = {10.48550/arXiv.2208.04933}
}

@inproceedings{gu2023mamba,
  author    = {A. Gu and T. Dao},
  title     = {Mamba: Linear-Time Sequence Modeling with Selective State Spaces},
  booktitle = {Conference on Language Modeling (COLM)},
  year      = {2024},
  doi       = {10.48550/arXiv.2312.00752}
}

@article{hopfield1982neural,
  author  = {J. J. Hopfield},
  title   = {Neural Networks and Physical Systems with Emergent Collective Computational Abilities},
  journal = {Proceedings of the National Academy of Sciences},
  year    = {1982},
  volume  = {79},
  number  = {8},
  pages   = {2554--2558},
  doi     = {10.1073/pnas.79.8.2554}
}

@inproceedings{ba2020using,
  author    = {J. Ba and G. Hinton and V. Mnih and J. Z. Leibo and C. Ionescu},
  title     = {Using Fast Weights to Attend to the Recent Past},
  booktitle = {Advances in Neural Information Processing Systems},
  volume    = {29},
  pages     = {4331--4339},
  publisher = {Curran Associates, Inc.},
  year      = {2016},
  doi       = {10.48550/arXiv.1610.06258}
}

@inproceedings{schlag2021linear,
  author    = {I. Schlag and K. Irie and J. Schmidhuber},
  title     = {Linear Transformers Are Secretly Fast Weight Programmers},
  booktitle = {Proceedings of the 38th International Conference on Machine Learning (ICML)},
  series    = {Proceedings of Machine Learning Research},
  volume    = {139},
  pages     = {9355--9366},
  publisher = {PMLR},
  year      = {2021}
}

@inproceedings{ramsauer2021hopfield,
  author    = {H. Ramsauer and B. Sch\"afl and J. Lehner and P. Seidl and M. Widrich and T. Adler and L. Gruber and M. Holzleitner and M. Pavlovi\'c and G. K. Sandve and V. Greiff and D. Kreil and M. Kopp and G. Klambauer and J. Brandstetter and S. Hochreiter},
  title     = {Hopfield Networks Is All You Need},
  booktitle = {9th International Conference on Learning Representations (ICLR)},
  publisher = {OpenReview.net},
  year      = {2021},
  doi       = {10.48550/arXiv.2008.02217}
}

@article{hyndman2011optimal,
  author  = {R. J. Hyndman and R. A. Ahmed and G. Athanasopoulos and H. L. Shang},
  title   = {Optimal Combination Forecasts for Hierarchical Time Series},
  journal = {Computational Statistics \& Data Analysis},
  year    = {2011},
  volume  = {55},
  number  = {9},
  pages   = {2579--2589},
  doi     = {10.1016/j.csda.2011.03.006}
}

@inproceedings{oreshkin2020nbeats,
  author    = {B. N. Oreshkin and D. Carpov and N. Chapados and Y. Bengio},
  title     = {N-BEATS: Neural Basis Expansion Analysis for Interpretable Time Series Forecasting},
  booktitle = {8th International Conference on Learning Representations (ICLR)},
  publisher = {OpenReview.net},
  year      = {2020},
  doi       = {10.48550/arXiv.1905.10437}
}

@article{salinas2020deepar,
  author  = {D. Salinas and V. Flunkert and J. Gasthaus and T. Januschowski},
  title   = {DeepAR: Probabilistic Forecasting with Autoregressive Recurrent Networks},
  journal = {International Journal of Forecasting},
  year    = {2020},
  volume  = {36},
  number  = {3},
  pages   = {1181--1191},
  doi     = {10.1016/j.ijforecast.2019.07.001}
}

@article{lim2021temporal,
  author  = {B. Lim and S. O. Arik and N. Loeff and T. Pfister},
  title   = {Temporal Fusion Transformers for Interpretable Multi-Horizon Time Series Forecasting},
  journal = {International Journal of Forecasting},
  year    = {2021},
  volume  = {37},
  number  = {4},
  pages   = {1748--1764},
  doi     = {10.1016/j.ijforecast.2021.03.012}
}

@inproceedings{rubanova2019latent,
  author    = {Y. Rubanova and R. T. Q. Chen and D. K. Duvenaud},
  title     = {Latent ODEs for Irregularly-Sampled Time Series},
  booktitle = {Advances in Neural Information Processing Systems},
  volume    = {32},
  pages     = {5320--5330},
  publisher = {Curran Associates, Inc.},
  year      = {2019},
  doi       = {10.48550/arXiv.1907.03907}
}

@inproceedings{kidger2020neural,
  author    = {P. Kidger and J. Morrill and J. Foster and T. J. Lyons},
  title     = {Neural Controlled Differential Equations for Irregular Time Series},
  booktitle = {Advances in Neural Information Processing Systems},
  volume    = {33},
  pages     = {6696--6707},
  publisher = {Curran Associates, Inc.},
  year      = {2020},
  doi       = {10.48550/arXiv.2005.08926}
}

@article{oord2016wavenet,
  author  = {A. van den Oord and S. Dieleman and H. Zen and K. Simonyan and O. Vinyals and A. Graves and N. Kalchbrenner and A. Senior and K. Kavukcuoglu},
  title   = {WaveNet: A Generative Model for Raw Audio},
  journal = {arXiv preprint arXiv:1609.03499},
  year    = {2016},
  doi     = {10.48550/arXiv.1609.03499}
}

@inproceedings{chollet2017xception,
  author    = {F. Chollet},
  title     = {Xception: Deep Learning with Depthwise Separable Convolutions},
  booktitle = {Proceedings of the IEEE Conference on Computer Vision and Pattern Recognition (CVPR)},
  pages     = {1800--1807},
  publisher = {IEEE},
  year      = {2017},
  doi       = {10.1109/CVPR.2017.195}
}

@inproceedings{glorot2010understanding,
  author    = {X. Glorot and Y. Bengio},
  title     = {Understanding the Difficulty of Training Deep Feedforward Neural Networks},
  booktitle = {Proceedings of the 13th International Conference on Artificial Intelligence and Statistics (AISTATS)},
  series    = {Proceedings of Machine Learning Research},
  volume    = {9},
  pages     = {249--256},
  publisher = {PMLR},
  year      = {2010}
}

@inproceedings{he2015delving,
  author    = {K. He and X. Zhang and S. Ren and J. Sun},
  title     = {Delving Deep into Rectifiers: Surpassing Human-Level Performance on ImageNet Classification},
  booktitle = {Proceedings of the IEEE International Conference on Computer Vision (ICCV)},
  pages     = {1026--1034},
  publisher = {IEEE},
  year      = {2015},
  doi       = {10.1109/ICCV.2015.123}
}

@article{ba2016layer,
  author  = {J. L. Ba and J. R. Kiros and G. E. Hinton},
  title   = {Layer Normalization},
  journal = {arXiv preprint arXiv:1607.06450},
  year    = {2016},
  doi     = {10.48550/arXiv.1607.06450}
}

@article{hendrycks2016gaussian,
  author  = {D. Hendrycks and K. Gimpel},
  title   = {Gaussian Error Linear Units (GELUs)},
  journal = {arXiv preprint arXiv:1606.08415},
  year    = {2016},
  doi     = {10.48550/arXiv.1606.08415}
}

@article{srivastava2014dropout,
  author  = {N. Srivastava and G. Hinton and A. Krizhevsky and I. Sutskever and R. Salakhutdinov},
  title   = {Dropout: A Simple Way to Prevent Neural Networks from Overfitting},
  journal = {Journal of Machine Learning Research},
  year    = {2014},
  volume  = {15},
  number  = {1},
  pages   = {1929--1958},
  doi     = {10.5555/2627435.2670313}
}

@inproceedings{loshchilov2019decoupled,
  author    = {I. Loshchilov and F. Hutter},
  title     = {Decoupled Weight Decay Regularization},
  booktitle = {7th International Conference on Learning Representations (ICLR)},
  publisher = {OpenReview.net},
  year      = {2019},
  doi       = {10.48550/arXiv.1711.05101}
}

@inproceedings{loshchilov2017sgdr,
  author    = {I. Loshchilov and F. Hutter},
  title     = {SGDR: Stochastic Gradient Descent with Warm Restarts},
  booktitle = {5th International Conference on Learning Representations (ICLR)},
  publisher = {OpenReview.net},
  year      = {2017},
  doi       = {10.48550/arXiv.1608.03983}
}

@inproceedings{pascanu2013difficulty,
  author    = {R. Pascanu and T. Mikolov and Y. Bengio},
  title     = {On the Difficulty of Training Recurrent Neural Networks},
  booktitle = {Proceedings of the 30th International Conference on Machine Learning (ICML)},
  series    = {Proceedings of Machine Learning Research},
  volume    = {28},
  pages     = {1310--1318},
  publisher = {PMLR},
  year      = {2013}
}

@article{williams1990efficient,
  author  = {R. J. Williams and J. Peng},
  title   = {An Efficient Gradient-Based Algorithm for On-Line Training of Recurrent Network Trajectories},
  journal = {Neural Computation},
  year    = {1990},
  volume  = {2},
  number  = {4},
  pages   = {490--501},
  doi     = {10.1162/neco.1990.2.4.490}
}

@article{Saidd2026Benchmark,
  author        = {Saidd, Nabeel Ahmad},
  title         = {A Controlled Comparison of Deep Learning
                   Architectures for Multi-Horizon Financial
                   Forecasting: Evidence from 918 Experiments},
  journal       = {arXiv preprint arXiv:2603.16886},
  year          = {2026},
  doi           = {10.48550/arXiv.2603.16886},
  eprint        = {2603.16886},
  archivePrefix = {arXiv},
  primaryClass  = {q-fin.ST}
}

@article{Saidd2026RNAD,
  author  = {Saidd, Nabeel Ahmad},
  title   = {RNAD: A Regime-Normalized, Tail-Robust
             and Direction-Aware Loss for Deep
             Time-Series Forecasting},
  year    = {2026},
  doi     = {10.2139/ssrn.7177678}
}

@article{Saidd2026Indicators,
  author  = {Saidd, Nabeel Ahmad},
  title   = {Do Technical Indicators Improve Deep
             Learning Forecasts? An Empirical
             Ablation Study Across Asset Classes},
  journal = {Preprints.org},
  year    = {2026},
  doi     = {10.20944/preprints202602.0692.v1}
}

@inproceedings{micikevicius2018mixed,
  author    = {P. Micikevicius and S. Narang and J. Alben and G. Diamos and E. Elsen and D. Garcia and B. Ginsburg and M. Houston and O. Kuchaiev and G. Venkatesh and H. Wu},
  title     = {Mixed Precision Training},
  booktitle = {6th International Conference on Learning Representations (ICLR)},
  publisher = {OpenReview.net},
  year      = {2018},
  doi       = {10.48550/arXiv.1710.03740}
}

\appendix

\clearpage
\section{Technical and Reproducibility Appendix}
\label{app:technical}
This appendix collects material that supports reproduction and technical scrutiny but would interrupt the main argument: the event-transition derivation underlying \cref{eq:maskedtransition,eq:batchedtransition}, tensor shapes, the ordered per-event algorithm, the configuration inventory across the three training scripts, and the per-seed metrics (\Cref{app:notation,app:shapes,app:ordering,app:repro,app:per-seed}). All values are configuration values or values already reported in the main text, and no new experiments are introduced.

\subsection{Event-transition derivation}
\label{app:notation}
\Cref{sec:problem} states the masked transition in terms of the scalar update indicator $u_\tau^k$ (\cref{eq:maskedtransition}) and in the batched form actually implemented (\cref{eq:batchedtransition}). The equivalent case-based form is closer to the code and makes explicit that masking selects between two already-computed branches rather than modifying either. Let $\bar t_\tau^k=\max\{t:t\leq\tau,\ t\text{ is a completed level-}k\text{ timestamp}\}$ be the no-later-than selection rule and $u_\tau^k=\mathbf{1}\{\bar t_\tau^k=\tau\}$ the resulting indicator of \cref{eq:update}. For a candidate transition $\Phi_k$,
\begin{equation}
 (r_{\tau+1}^k,M_{\tau+1}^k)=
 \begin{cases}
 \Phi_k(r_\tau^k,M_\tau^k,X_\tau^k),&u_\tau^k=1,\\
 (r_\tau^k,M_\tau^k),&u_\tau^k=0,
 \end{cases}
\end{equation}
which coincides with \cref{eq:maskedtransition} because $u_\tau^k\in\{0,1\}$, so the convex combination always selects exactly one branch. In the batched implementation the same selection is applied per sample: entry $b$ of the mask vector $m_\tau^k$ (\cref{eq:batchedmask}) is the scalar indicator of stream $b$, and \cref{eq:batchedtransition} broadcasts it over the state and memory dimensions. The base level has $u_\tau^1\equiv1$, since the anchor stream is defined by level-1 bar completions. This establishes preservation of persistent state, not computational skipping: candidate encodings are formed for all supplied windows regardless of the mask (\Cref{sec:streaming}).

The causal contract of \cref{eq:causal} is event-level. The lower-level read is appended to its history before the higher-level evidence readout (\cref{eq:evidence}), so no future \emph{event} is used, but a higher-level update can consume a lower-level representation produced at the same event. This is the reason for the qualified entry~9 of \cref{tab:audit}, and it is the one place in the specification where ``causal'' must be read at event granularity rather than at the granularity of individual computation steps within an event.

\subsection{Tensor shapes across the pipeline}
\label{app:shapes}
\Cref{tab:shapes} traces the tensor shape at each stage of \Cref{sec:architecture} for the default configuration ($d_s=72$, $d_m=18$, window length $T=32$, rank $d_r=36$).

\begin{table}[!htbp]\centering\small
\caption{Tensor shapes at each pipeline stage in the default configuration ($d_s=72$, $d_m=18$, $T=32$, $d_r=36$). $B$ is the batch size and $L$ the number of configured levels for the asset in question.}
\label{tab:shapes}
\begin{tabular}{>{\raggedright\arraybackslash}p{4.6cm}>{\raggedright\arraybackslash}p{7.8cm}}
\toprule Stage & Shape \\\midrule
Raw window $X^k$ & $B\times T\times5$ \\
Derived features $d_t^k$ & $B\times T\times6$ \\
Encoder output $e^k$ (\cref{eq:encoder}) & $B\times72$ \\
Update mask $m_\tau^k$ (\cref{eq:batchedmask}) & $B$ \\
Memory input $x_t$ (base / higher level) & $B\times144$ / $B\times216$ \\
Memory $M_t$ & $B\times72\times18$ \\
State $r_t$ & $B\times72$ \\
Resonance stack $R$ & $B\times L\times72$ \\
Evidence readout $E_i$ (\cref{eq:evidence}) & $B\times72$ \\
Head input (\cref{eq:head}) & $B\times150$ \\
Forecast $\hat y_{t+1}^k$ & $B\times1$ \\
\bottomrule
\end{tabular}
\end{table}

\FloatBarrier

At the base level, the memory input concatenates a 72-dimensional encoder output with a 72-dimensional prior resonance context. Higher levels additionally receive 72-dimensional bottom-up evidence, which accounts for the 144- versus 216-dimensional memory input of \Cref{sec:memory}. The read $\bar v=Mk$ and residual $s=v-\bar v$ enter \cref{eq:memory,eq:state} as described there. The shapes make explicit that the association store is a $72\times18$ matrix per sample, not a 72-dimensional vector state, which is the structural difference from a standard recurrent cell noted in \Cref{sec:related}.

\subsection{Ordered per-event algorithm}
\label{app:ordering}
The exact ordering of the per-event computation described in \Cref{sec:streaming} is what makes the causal qualification of \Cref{sec:problem} (entry~9 of \cref{tab:audit}) precise. It is restated here as an ordered procedure:
\begin{enumerate}[leftmargin=*]
\item select the latest completed window and update indicator for each level (\Cref{sec:problem});
\item encode all supplied windows and construct anchor features (\Cref{sec:architecture});
\item apply prior-state resonance (\cref{eq:resonance});
\item update the base level, which is always active;
\item append each current lower-level read and compute the adjacent higher-level evidence (\cref{eq:evidence});
\item compute candidate higher-level transitions and apply their masks (\cref{eq:batchedtransition});
\item apply posterior resonance to the resulting states (\cref{eq:resonance}); and
\item emit one BPS forecast per configured level (\cref{eq:head}).
\end{enumerate}
Steps 3 and 7 are the two resonance applications of \Cref{sec:architecture}, and step 5 is the ordered evidence computation responsible for the qualified entry~9. Price reconstruction occurs only after the BPS forecast is emitted in step 8, consistent with the network being optimized purely in BPS space.

\subsection{Configuration inventory across the training scripts}
\label{app:repro}
The production trainer, the reported configuration table, and the ablation trainer do not share identical defaults (\Cref{sec:streaming,sec:provenance}), and these artifact-level configurations must not be silently merged. \Cref{tab:configinventory} consolidates the values referenced individually in the main text.

\begin{table}[!htbp]
\centering
\small
\caption{Key training settings of the production trainer, the reported main-results recipe, and the ablation trainer. The configurations differ only in the seed lists and in the ablation epoch budget; shared values are shown explicitly rather than inferred.}
\label{tab:configinventory}

\begin{tabular}{>{\raggedright\arraybackslash}p{3.2cm}
                >{\raggedright\arraybackslash}p{3.0cm}
                >{\raggedright\arraybackslash}p{3.0cm}
                >{\raggedright\arraybackslash}p{3.0cm}}
\toprule
Setting & Production & Main results & Ablations \\
\midrule
Batch size & 256 & 256 & 256 \\
Max epochs & 50 & 50 & 5 \\
Evaluation streams & 8 & 8 & 8 \\
Optimizer & AdamW & AdamW & AdamW \\
Learning rate & $10^{-3}$ & $10^{-3}$ & $10^{-3}$ \\
Weight decay & $10^{-5}$ & $10^{-5}$ & $10^{-5}$ \\
Gradient clipping & 1.0 & 1.0 & 1.0 \\
Patience & 15 & 15 & 15 \\
Schedule & Cosine & Cosine & Cosine \\
Seeds & 42 & 42, 151, 359 & 42 \\
Precision & CUDA AMP & CUDA AMP & CUDA AMP \\
\bottomrule
\end{tabular}
\end{table}

One discrepancy should be resolved from the available configuration and checkpoint metadata: the reported recipe lists eight-stream validation and test, whereas the standalone evaluator uses one stream and reconstructs missing checkpoint configuration fields from defaults and tensor shapes. Because the production trainer writes seed-specific checkpoints and the dataset configuration is available in the repository, the reported three-seed aggregates can be regenerated from the repository together with a compatible vendor feed. Per-seed aggregate metrics and seed-specific checkpoints are available, and per-event outputs can be generated from those checkpoints. No target-space metric is emitted by the evaluators, and no persistence baseline, confidence interval, or paired test is provided (\Cref{sec:results}). The ablation variants A1--A3 are interface-compatible interventions, not capacity-controlled counterfactuals (\Cref{sec:ablation}).

\subsection{Per-seed metrics}
\label{app:per-seed}
The per-seed aggregate metrics reported here are for seeds 42, 151, and 359. They are aggregate metrics, not per-event predictions. \Cref{tab:harn-per-seed} lists HARN's per-seed metrics for each asset and timeframe, and \Cref{tab:baseline-per-seed} lists those of PatchTST and TimeXer. The baseline per-seed CSV does not contain a timeframe column, so the timeframe labels in \Cref{tab:baseline-per-seed} are inferred from \Cref{tab:mainresults,tab:fairness}. The mean $\pm$ standard deviation entries in \Cref{tab:mainresults} are computed from these per-seed values, and the reported standard deviations follow the population convention, in which the sum of squared deviations is divided by the number of seeds. Recomputing the means and standard deviations from \Cref{tab:harn-per-seed,tab:baseline-per-seed} reproduces the corresponding entries in \Cref{tab:mainresults}, and no discrepancies were found. Per-seed training trajectories and per-event predictions are not included as precomputed files and can be generated from the checkpoints and preprocessing pipeline.

\begin{longtable}{llr
    S[table-format=1.6]
    S[table-format=1.6]
    S[table-format=1.6]
    S[table-format=2.6]}
\caption{HARN per-seed aggregate test metrics for seeds 42, 151, and 359, grouped by asset and timeframe. The means and standard deviations in \Cref{tab:mainresults} are computed from these values.}
\label{tab:harn-per-seed}\\
\toprule
\textbf{Asset} & \textbf{TF} & \textbf{Seed} &
\textbf{MAE} & \textbf{RMSE} & \textbf{sMAPE} & \textbf{MASE} \\
\midrule
\endfirsthead

\toprule
\textbf{Asset} & \textbf{TF} & \textbf{Seed} &
\textbf{MAE} & \textbf{RMSE} & \textbf{sMAPE} & \textbf{MASE} \\
\midrule
\endhead

\bottomrule
\endfoot

\multirow{6}{*}{AAPL}
& \multirow{3}{*}{H1}
& 42  & 1.058752 & 1.801058 & 0.425964 & 4.815850 \\
& & 151 & 1.060329 & 1.800236 & 0.426467 & 4.823021 \\
& & 359 & 1.058117 & 1.799379 & 0.425760 & 4.812961 \\
\cmidrule(lr){2-7}
& \multirow{3}{*}{M15}
& 42  & 0.506980 & 0.933028 & 0.204653 & 1.325713 \\
& & 151 & 0.507987 & 0.933396 & 0.205064 & 1.328345 \\
& & 359 & 0.508120 & 0.934044 & 0.205123 & 1.328694 \\

\midrule

\multirow{9}{*}{EURUSD}
& \multirow{3}{*}{H1}
& 42  & 0.000617 & 0.000907 & 0.053021 & 3.468855 \\
& & 151 & 0.000618 & 0.000908 & 0.053103 & 3.474228 \\
& & 359 & 0.000613 & 0.000904 & 0.052670 & 3.445965 \\
\cmidrule(lr){2-7}
& \multirow{3}{*}{H4}
& 42  & 0.001227 & 0.001746 & 0.105450 & 13.545654 \\
& & 151 & 0.001225 & 0.001743 & 0.105267 & 13.522090 \\
& & 359 & 0.001229 & 0.001746 & 0.105576 & 13.562517 \\
\cmidrule(lr){2-7}
& \multirow{3}{*}{M15}
& 42  & 0.000305 & 0.000465 & 0.026173 & 0.861150 \\
& & 151 & 0.000309 & 0.000469 & 0.026576 & 0.874469 \\
& & 359 & 0.000305 & 0.000466 & 0.026182 & 0.861444 \\

\midrule

\multirow{9}{*}{USDCHF}
& \multirow{3}{*}{H1}
& 42  & 0.000527 & 0.000762 & 0.066730 & 3.237441 \\
& & 151 & 0.000525 & 0.000763 & 0.066459 & 3.224394 \\
& & 359 & 0.000527 & 0.000763 & 0.066706 & 3.236370 \\
\cmidrule(lr){2-7}
& \multirow{3}{*}{M15}
& 42  & 0.000263 & 0.000391 & 0.033288 & 0.800227 \\
& & 151 & 0.000263 & 0.000391 & 0.033265 & 0.799668 \\
& & 359 & 0.000263 & 0.000391 & 0.033299 & 0.800498 \\
\cmidrule(lr){2-7}
& \multirow{3}{*}{M30}
& 42  & 0.000374 & 0.000544 & 0.047287 & 1.610327 \\
& & 151 & 0.000372 & 0.000543 & 0.047128 & 1.604894 \\
& & 359 & 0.000373 & 0.000543 & 0.047209 & 1.607693 \\

\midrule

\multirow{6}{*}{XAUUSD}
& \multirow{3}{*}{M15}
& 42  & 4.532421 & 6.820130 & 0.108462 & 3.485143 \\
& & 151 & 4.523920 & 6.815818 & 0.108263 & 3.478606 \\
& & 359 & 4.530076 & 6.826767 & 0.108404 & 3.483340 \\
\cmidrule(lr){2-7}
& \multirow{3}{*}{M5}
& 42  & 2.550012 & 3.844527 & 0.061048 & 1.135077 \\
& & 151 & 2.553465 & 3.845944 & 0.061131 & 1.136614 \\
& & 359 & 2.552857 & 3.842431 & 0.061117 & 1.136344 \\

\end{longtable}

\begin{longtable}{lll r
    S[table-format=2.6]
    S[table-format=2.6]
    S[table-format=1.6]
    S[table-format=2.6]}
\caption{Per-seed aggregate test metrics for PatchTST and TimeXer, grouped by asset, model, and timeframe. Timeframe labels are inferred from \Cref{tab:mainresults,tab:fairness} because the source CSV has no timeframe column.}
\label{tab:baseline-per-seed}\\
\toprule
\textbf{Asset} & \textbf{Model} & \textbf{TF} & \textbf{Seed} &
\textbf{MAE} & \textbf{RMSE} & \textbf{sMAPE} & \textbf{MASE} \\
\midrule
\endfirsthead

\toprule
\textbf{Asset} & \textbf{Model} & \textbf{TF} & \textbf{Seed} &
\textbf{MAE} & \textbf{RMSE} & \textbf{sMAPE} & \textbf{MASE} \\
\midrule
\endhead

\bottomrule
\endfoot

\multirow{6}{*}{AAPL}
& \multirow{3}{*}{PatchTST} & \multirow{3}{*}{M15}
& 42  & 1.358967 & 1.949987 & 0.527755 & 3.553591 \\
& & & 151 & 2.095496 & 2.865226 & 0.791903 & 5.479553 \\
& & & 359 & 2.768740 & 3.291260 & 1.064283 & 7.240032 \\
\cmidrule(lr){2-8}
& \multirow{3}{*}{TimeXer} & \multirow{3}{*}{M15}
& 42  & 3.855025 & 4.600736 & 1.480743 & 10.080581 \\
& & & 151 & 1.868187 & 2.529758 & 0.764724 & 4.885158 \\
& & & 359 & 2.922446 & 3.709866 & 1.100429 & 7.641962 \\

\midrule

\multirow{6}{*}{EURUSD}
& \multirow{3}{*}{PatchTST} & \multirow{3}{*}{M15}
& 42  & 0.000658 & 0.000837 & 0.056428 & 1.859308 \\
& & & 151 & 0.001135 & 0.001388 & 0.097050 & 3.206570 \\
& & & 359 & 0.000484 & 0.000676 & 0.041554 & 1.366752 \\
\cmidrule(lr){2-8}
& \multirow{3}{*}{TimeXer} & \multirow{3}{*}{M15}
& 42  & 0.011227 & 0.013710 & 0.963514 & 31.726979 \\
& & & 151 & 0.010385 & 0.012435 & 0.891052 & 29.348186 \\
& & & 359 & 0.010901 & 0.013006 & 0.935528 & 30.805515 \\

\midrule

\multirow{6}{*}{USDCHF}
& \multirow{3}{*}{PatchTST} & \multirow{3}{*}{M15}
& 42  & 0.000780 & 0.001012 & 0.099158 & 2.371872 \\
& & & 151 & 0.000571 & 0.000773 & 0.072218 & 1.738637 \\
& & & 359 & 0.000563 & 0.000805 & 0.071415 & 1.712944 \\
\cmidrule(lr){2-8}
& \multirow{3}{*}{TimeXer} & \multirow{3}{*}{M15}
& 42  & 0.014517 & 0.015962 & 1.828844 & 44.169506 \\
& & & 151 & 0.013430 & 0.014761 & 1.693373 & 40.862840 \\
& & & 359 & 0.014486 & 0.016231 & 1.825691 & 44.073778 \\

\midrule

\multirow{6}{*}{XAUUSD}
& \multirow{3}{*}{PatchTST} & \multirow{3}{*}{M5}
& 42  & 8.634885 & 11.674337 & 0.205627 & 3.843614 \\
& & & 151 & 11.015642 & 13.945930 & 0.264041 & 4.903351 \\
& & & 359 & 9.039866 & 12.415752 & 0.215399 & 4.023882 \\
\cmidrule(lr){2-8}
& \multirow{3}{*}{TimeXer} & \multirow{3}{*}{M5}
& 42  & 16.241031 & 21.248008 & 0.380107 & 7.229309 \\
& & & 151 & 16.801729 & 21.078022 & 0.395585 & 7.478891 \\
& & & 359 & 12.849670 & 14.958773 & 0.308100 & 5.719726 \\

\end{longtable}

\end{document}